%% file: flaml.tex
\documentclass{article}

\usepackage{microtype}
\usepackage{graphicx}
\usepackage{subcaption}
\usepackage{booktabs} 

\usepackage{hyperref}
\usepackage{amsmath,amssymb}
\usepackage{xspace}
\usepackage{ifthen}
\usepackage{listings}
\usepackage{xcolor}
\definecolor{codegreen}{rgb}{0,0.6,0}
\definecolor{codegray}{rgb}{0.5,0.5,0.5}
\definecolor{codepurple}{rgb}{0.58,0,0.82}
\definecolor{backcolour}{rgb}{0.95,0.95,0.92}
\lstdefinestyle{mystyle}{
    backgroundcolor=\color{backcolour},   
    commentstyle=\color{codegreen},
    keywordstyle=\color{magenta},
    numberstyle=\tiny\color{codegray},
    stringstyle=\color{codepurple},
    basicstyle=\ttfamily\footnotesize,
    breakatwhitespace=false,         
    breaklines=true,                 
    captionpos=b,                    
    keepspaces=true,                 
    numbers=left,                    
    numbersep=5pt,                  
    showspaces=false,                
    showstringspaces=false,
    showtabs=false,                  
    tabsize=2
}

\newcommand{\compilehidecomments}{false}
\ifthenelse{ \equal{\compilehidecomments}{true} }{%
\newcommand{\Outline}[1]{{\color{blue}{\text{} #1}}}
\newcommand{\note}[1]{{\color{red}{\text{} #1}}}
}{
	\newcommand{\Outline}[1]{}
	\newcommand{\note}[1]{}
}

\def \FLAML {FLAML\xspace}

\def \bp {\mathbf{h}}
\def \bu {\mathbf{u}}

\newtheorem{observation}{Observation}
\newtheorem{property}{Property}
\usepackage[accepted]{arxiv}

\mlsystitlerunning{\FLAML: A Fast and Lightweight AutoML Library}

\begin{document}

\twocolumn[
\mlsystitle{\FLAML: A Fast and Lightweight AutoML Library}




\begin{mlsysauthorlist}
\mlsysauthor{Chi Wang}{ms}
\mlsysauthor{Qingyun Wu}{ms}
\mlsysauthor{Markus Weimer}{ms}
\mlsysauthor{Erkang Zhu}{ms}
\end{mlsysauthorlist}

\mlsysaffiliation{ms}{Microsoft Corporation, Redmond, WA, USA}

\mlsyscorrespondingauthor{Chi Wang}{wang.chi@microsoft.com}

\mlsyskeywords{Automated Machine Learning, Computational Cost}

\vskip 0.3in

\begin{abstract}
We study the problem of using low computational cost to automate the choices of learners and hyperparameters for an ad-hoc training dataset and error metric, by conducting trials of different configurations on the given training data. We investigate the joint impact of multiple factors on both trial cost and model error, and propose several design guidelines. 
Following them, we build a fast and lightweight library FLAML which optimizes for low computational resource in finding accurate models. FLAML integrates several simple but effective search strategies into an adaptive system. It significantly outperforms top-ranked AutoML libraries on a large open source AutoML benchmark under equal, or sometimes orders of magnitude smaller budget constraints.
\end{abstract}

]



\printAffiliationsAndNotice{}  

\input{sec_intro}

\input{sec_related_work}
\input{sec_problem}
\input{sec_system}

\input{sec_exp}

\input{sec_conclusion}

\section*{Acknowledgments}
The authors appreciate suggestions from Surajit Chaudhuri, Nadiia Chepurko, Alex Deng, Anshuman Dutt, Johannes Gehrke, Silu Huang, Christian Konig, and Haozhe Zhang.

\bibliography{flaml}
\bibliographystyle{arxiv}

\appendix
\input{sec_appendix}

\end{document}

%% file: sec_intro.tex
\section{Introduction}
\begin{figure*}[h]
  \centering
  \begin{subfigure}{0.32\textwidth}
\includegraphics[width=\columnwidth]{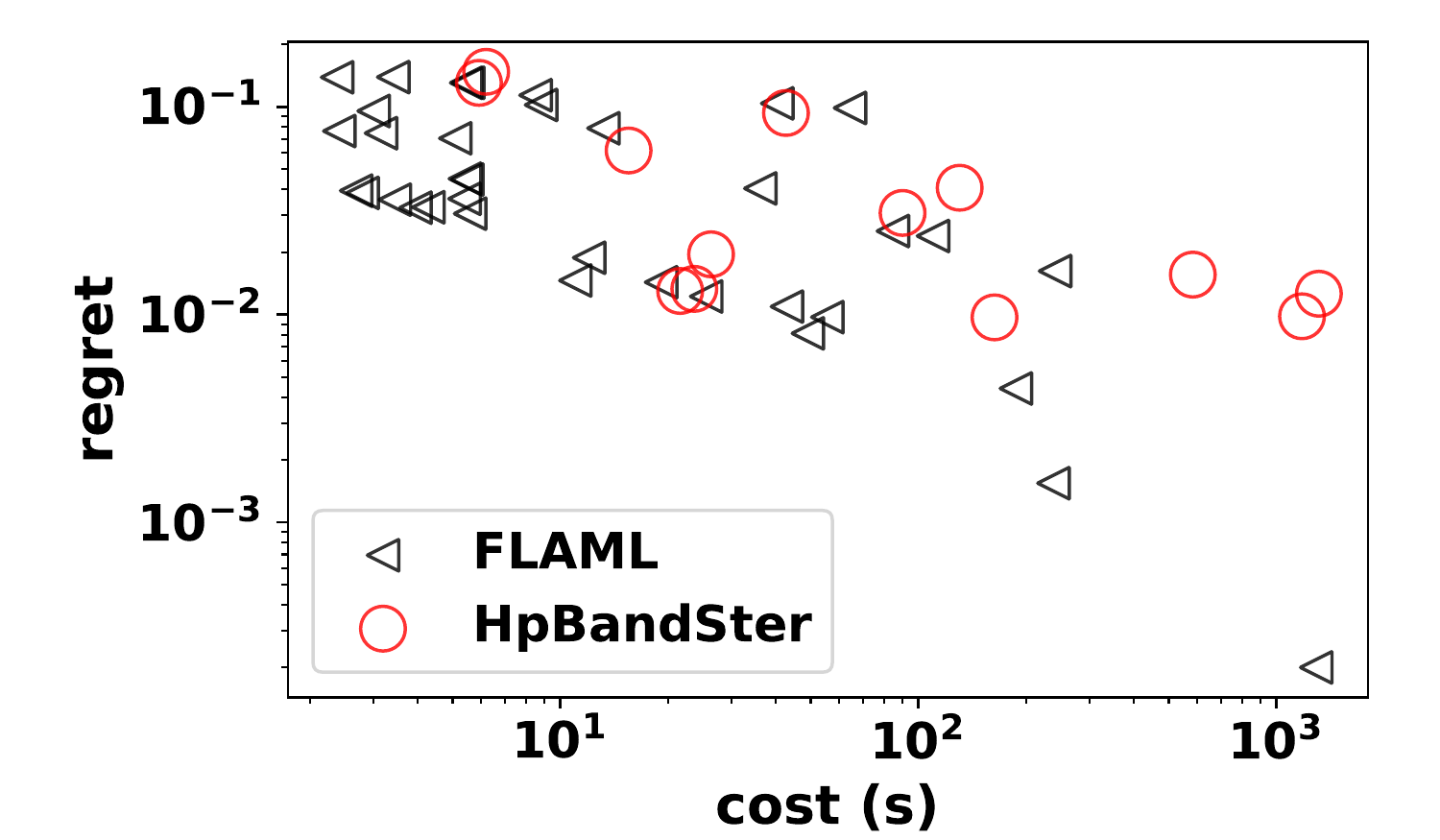}%
\caption{Model auc regret vs. training cost for every trial}%
\end{subfigure}\hfill%
\begin{subfigure}{0.32\textwidth}
\includegraphics[width=\columnwidth]{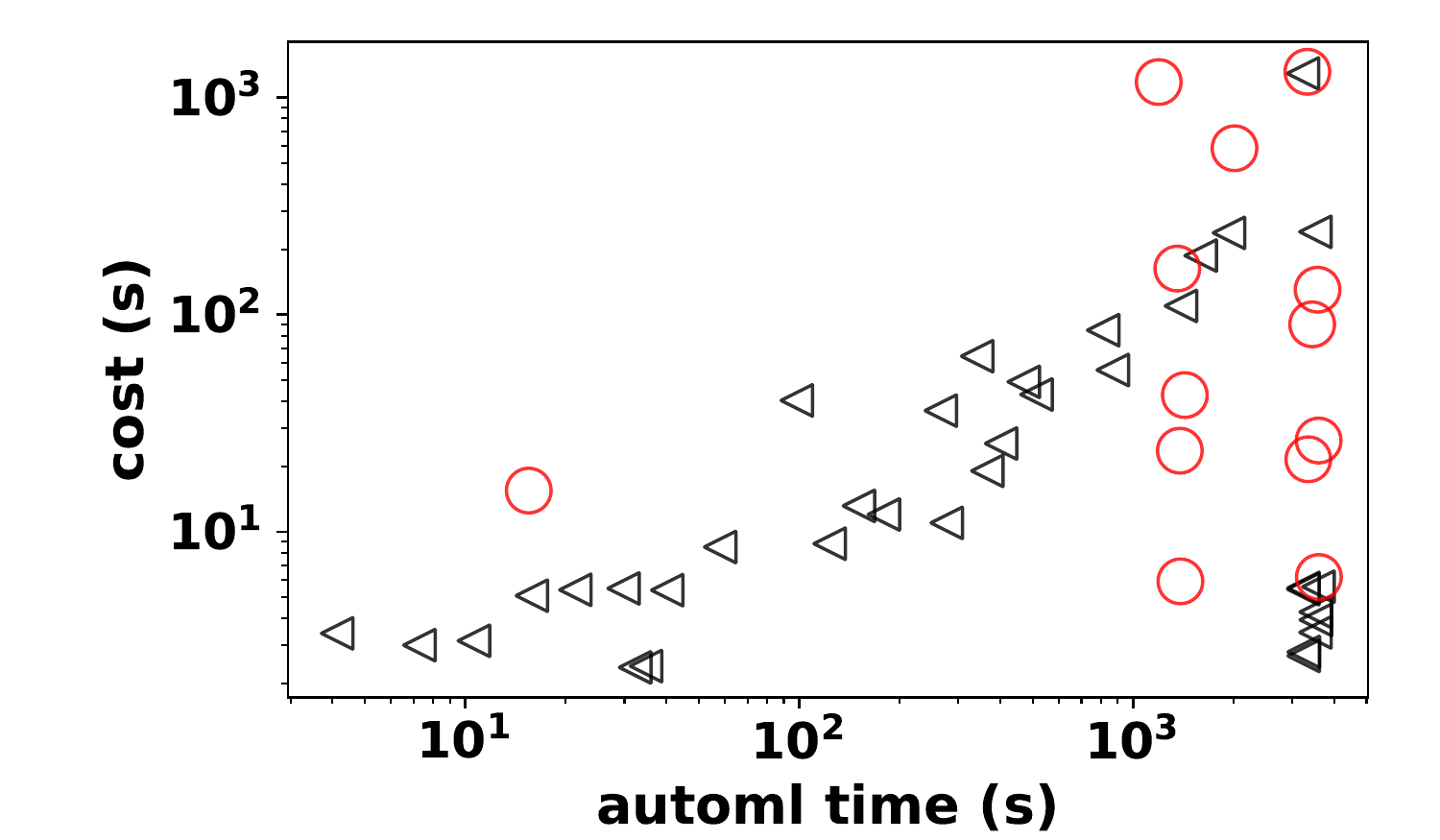}%
\caption{Trial cost vs. the total time from start when each trial is finished }%
\end{subfigure}\hfill%
\begin{subfigure}{0.32\textwidth}
\includegraphics[width=\columnwidth]{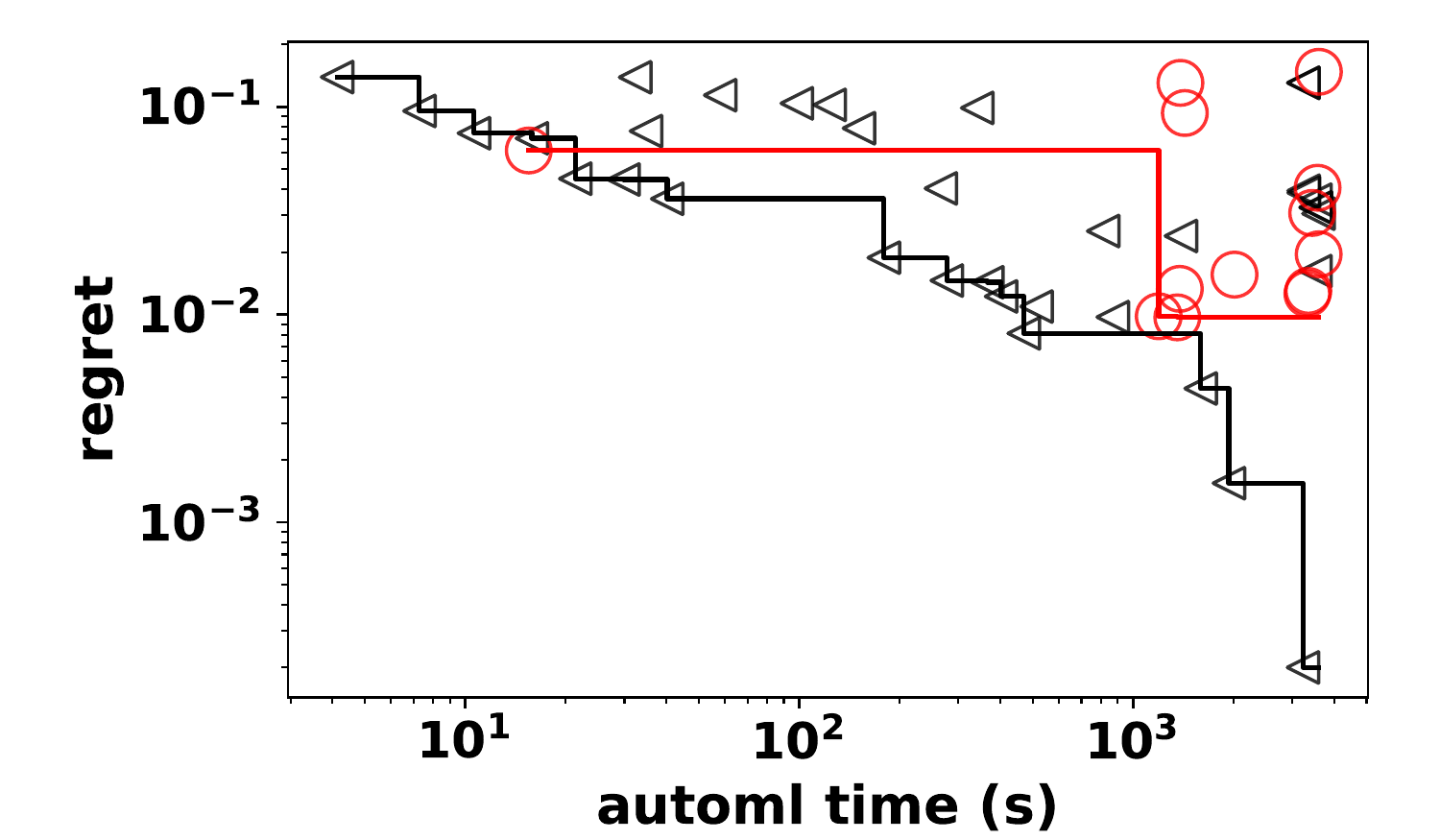}%
\caption{Model auc regret vs. the total time from start when each trial is finished}%
\end{subfigure}\hfill%
  \caption{Example of search performance for \FLAML vs. a baseline in the same search space.
  Each marker corresponds to one trial of configuration evaluation in a particular method. Model auc regret=best\_auc-model\_auc. Each marker corresponds to one trial of configuration evaluation in a particular method. Subfigure (a) suggests that 
  \FLAML makes fewer expensive trials with high error (top right corner) than HpBandSter. Subfigure (b) further displays that the expense of trials made by \FLAML grows gradually with total time spent, while for HpBandSter there is no such trend. As a result, subfigure (c) shows that \FLAML outperforms in both early and late stages. 
  }
  \label{fig:casestudy}
\end{figure*}

It is predicted that in the next 10 years, hundreds of thousands of small teams will build millions of ML-infused applications -- most just moderately remunerative, but with huge collective value~\cite{agrawal2020cloudy}. Operating by large teams of ML experts and running on massive dedicated infrastructures is not well justified for these new applications. 
That motivates fast and economical software solutions to Automated Machine Learning (AutoML): Given a training dataset and an error metric, use \emph{low computational cost} to search for learner and hyperparameter choices and produce models optimizing the error metric in short time. 

To provide a concrete context, let us consider the use of ML in database systems. The database community has grown an increasing interest
of integrating data-driven decision making components fueled by machine learning techniques.
For example, classification or regression models have been explored for indexing~\cite{kraska2018case,galakatos2019tree}, 
cardinality and selectivity estimation~\cite{kipf2019learned,Dutt2019selectivity}, query performance prediction~\cite{pvldb:MarcusP19}, and workload forecasting~\cite{ma2018query}.
These models make predictions by learning from a large amount of labeled data, which are generated automatically by the system for each dataset or workload instance. For example, a selectivity estimation model can be built for each table or join expression using selectivity labels generated from synthetic queries or a given workload~\cite{Dutt2019selectivity,dutt2020efficiently}, and the best model configurations vary per instance of training dataset. AutoML solutions for these applications are required to be fast and economic, as the system needs to select hyperparameters or learners frequently on different training data (for numerous tables, join expressions, and frequent updates), and continuously and timely deploy them~\cite{renggli2019continuous}. Computational resource of the system is precious (e.g., for answering database queries), and only a small fraction can be allocated to AutoML, e.g., a few CPU minutes per selectivity estimation model.

A number of AutoML libraries have been developed, 
which usually involve multiple trials of different configurations. 
One drawback in existing solutions is 
they require long time or large amounts of resources to produce accurate models for large scale training datasets.
For example, given one CPU hour, when tested on a recent large-scale AutoML benchmark~\cite{Gijsbers2019benchmark}, the state-of-the-art solutions underperform a tuned random forest baseline on 36-51\% of the tasks. And the ratio is even higher when the budget is smaller.

To address the problem systematically, it is desirable to factor the \emph{trial cost}, i.e., the CPU cost of training and assessing the model error, explicitly in the AutoML problem. We recognize that the cost of one trial is jointly decided by the following variables: the choice of learner, a subset of the hyperparameters for the chosen learner, the size of the training data, and the resampling strategy. 
Those variables also affect the \emph{trial error} (i.e., the assessed model error) jointly. 
Given an ad-hoc dataset, an AutoML solution that is only optimized for low trial error may invoke unnecessarily expensive trials, while a solution that is only optimized for low trial cost may keep making cheap but erroneous trials. 
Some hyperparameter optimization methods made an effort to balance the two objectives, but the scope is limited and most systems target resource-consuming clusters~\cite{snoek2012practical,ICLR:li2017hyperband,falkner2018,Liaw19,Li2020}. 
No previous AutoML system handles the complex dependency among the multiple variables mentioned above.
Though challenging, it is desired to have an economical system that holistically considers the multiple factors in the cost-error tradeoff, and handles different tasks robustly and efficiently.

We design and implement a lightweight Python library \FLAML\footnote{ \url{https://github.com/microsoft/FLAML}}.
\FLAML\ leverages the structure of the search space 
to choose a search order optimized for both cost and error. It iteratively decides the learner, hyperparameter, sample size and resampling strategy while leveraging their compound impact on both cost and error as the search proceeds. First, we analyze the relation of these factors and deduce desirable properties of an economical AutoML system. To satisfy these properties, we 
integrate several non-traditional search strategies judiciously because commonly employed strategies do not sufficiently exploit the analyzed relations of the multiple factors. 
Overall, the search tends to gradually move from cheap trials and inaccurate models to expensive trials and accurate models (a typical example is illustrated in Figure~\ref{fig:casestudy}). \FLAML\ is designed for robustly adapting to an ad-hoc dataset out of the box,
without relying on expensive preparation such as meta-learning. In fact, our system has almost no computational overhead beyond the trial cost of each configuration. 

We perform extensive evaluation using a recent open source AutoML benchmark~\cite{Gijsbers2019benchmark} plus regression datasets from a regression benchmark~\cite{Olson2017PMLB}.
With varying time budget from one minute to one hour, \FLAML\ outperforms top three open-source AutoML libraries as well as a commercial cloud-based AutoML service 
in a majority of the tasks given equal or smaller budget, with significant margins. 
We study an application to selectivity estimation in the end.

%% file: sec_related_work.tex
\section{Related Work} \label{sec:related_work}

First, we review the top-performing open-source AutoML libraries according to the AutoML Benchmark~\cite{Gijsbers2019benchmark}.  
(1) Auto-sklearn~\cite{feurer2015efficient} is declared the overall winner of the ChaLearn AutoML Challenge 1 in 2015-2016 and 2 in 2017-2018. It employs Bayesian optimization (BO)~\cite{hutter2011} for hyperparameter tuning and learner selection, and uses meta-learning to warm-start the search procedure with a few pipelines. (2) TPOT~\cite{Olson2016TPOT} (Tree-based Pipeline Optimization Tool) constructs machine learning pipelines of arbitrary length using scikit-learn learners and XGBoost and uses genetic programming for hyperparameter tuning. (3) H2O AutoML~\cite{H2O} is a Java-based library. It performs randomized grid search for each learner in the H2O machine learning package, in addition to XGBoost. The learners are ordered manually and each learner is allocated a predefined portion of search iterations. 
They all use model ensembles to boost accuracy.

A number of commercial platforms are available: Amazon AWS SageMaker~\cite{liberty2020elastic}, DataRobot, Google Cloud AutoML Tables, Microsoft AzureML AutoML, Salesforce TransmogrifAI, H2O Driverless AI, Darwin AutoML and Oracle AutoML. They provide end-to-end AutoML service, i.e., directly consuming uncleaned raw data and then producing trained models and predictions.  

To summarize the learnings from existing AutoML systems, the dominating approach is based on trials in a large search space. The order of the trials thus has a large impact in the search efficiency. Meta-learning is one technique often proposed to improve the search order, with the assumption that one can collect a large number of datasets and experiments for meta-training, and the performance of learners and hyperparameters from these experiments is indicative of their future performance in new datasets and tasks~\cite{feurer2015efficient,fusi2018probabilistic,Shang2019DDS}. In addition, ensemble of multiple learners is often considered useful for boosting accuracy at the cost of increased inference latency~\cite{agtabular}. 

\FLAML is designed to perform efficiently and robustly without relying on meta-learning or ensemble at first order, for several usability reasons. First, this makes \FLAML an easy plug-in in new application scenarios, without requiring a developer to collect many diverse meta-training datasets before being able to use it. Second, it allows the user to easily customize learners, search spaces and optimization metrics and use \FLAML immediately after the customization, without waiting for another expensive round of meta-learning if any of these changes. Third, 
our customers prefer single learners over ensembles due to the advantage in model complexity, inference latency, ease of deployment, debuggability and explanability. How to leverage meta-learning and ensemble with good usability is interesting future work.

One notable standalone subarea in AutoML is neural architecture search (NAS)~\cite{JMLR2019NAS} which specifically targets neural networks. 
Most application scenarios of NAS involve unstructured data like images and text. While the search space and application scenario are different, our design principles in cost minimization might be applicable.

%% file: sec_problem.tex
\section{API, Formulation and Analysis}\label{sec:probelm}
\FLAML is implemented in Python because of its popularity in data science. 
It has a scikit-learn~\cite{JMLR:Pedregosa2011} style API:
\begin{lstlisting}[language=Python]
from flaml import AutoML
automl = AutoML()
automl.fit(X_train, y_train, task='classification')
prediction = automl.predict(X_test)
\end{lstlisting}
Additional settings include time budget, optimization metric, estimator list etc. It is easy to add customized learners or metrics in \FLAML:
\begin{lstlisting}[language=Python]
# MyLearner is a custom estimator class
automl.add_learner(learner_name='mylearner', learner_class=MyLearner)
# mymetric is a custom metric function
automl.fit(X_train, y_train, metric=mymetric, time_budget=60,     estimator_list=['mylearner','xgboost'])
\end{lstlisting}

The main innovation of \FLAML is in its fit() method: automatically
producing
an accurate model (measured by a given error metric)
for an ad-hoc featurized dataset\footnote{Given existing fast automatic featurization libraries such as \emph{autofeat}~\cite{horn2019autofeat} and \emph{azureml-sdk}~\cite{mukunthu2019practical}, FLAML does not innovate on featurization techniques, though the system can easily support feature preprocessors.
}. 
\subsection{Formulation}
We consider $L$ learners, each with its own set of hyperparameters. The learners can be customized by users, as long as they have well-defined train and prediction methods and search space of hyperparameters.
We denote the search space of hyperparameters for learner $l$ as $H_l$. For each trial, we can choose a learner $l$, the hyperparameters $\bp\in H_l$, together with two other variables: sample size $s$ of training data, and resampling strategy $r$. $s$ is an integer to denote the number of examples in a sample of training data, and $r\in\{\text{cv},\text{holdout}\}$ is a binary choice between k-fold cross-validation and holdout with ratio $\rho$.\footnote{In general, we can consider a large search space for the resampling strategy by making $k$ and $\rho$ variables as well. We make $k$ and $\rho$ constants in this work to simplify the problem. 
}
A \textit{learning configuration} is defined as a tuple $\chi=(l,\bp,s,r)$. 
When we make a trial with $\chi$, 
we can obtain a validation error $\tilde{\epsilon}(\chi)$ and a model $M(\chi)$. 
Depending on whether the resampling strategy is cross validation or holdout, the model $M$ corresponds to training data of size $s$ or $s\cdot(1-\rho)$, where $\rho$ is the holdout ratio, and the error $\tilde\epsilon$ corresponds to cross validation error or error on the heldout validation data. $\tilde\epsilon$ is a proxy of the actual error $\epsilon(M)$ on unseen test data. 
The cost of the trial is mainly the CPU time of training and testing using cross-validation or holdout, denoted as $\kappa(\chi)$.  The goal of fast and economical AutoML is to minimize the total cost before finding a model with the lowest test error. The total cost is expected to increase as the test error decreases, and desired to be approximately optimal. 

\subsection{Analysis}\label{sec:analysis}
\begin{figure}
  \centering
        \includegraphics[width=\linewidth]{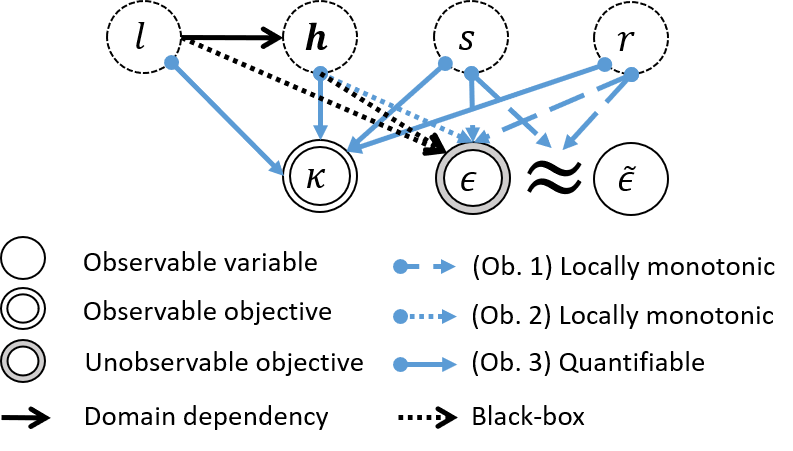}
    \caption{Relations among the variables.}\label{fig:relation}
\end{figure}

We first analyze the factors considered in our search sequence and several desirable properties of the search dynamics about them.
Figure~\ref{fig:relation} summarizes the relations among several variables, using notations summarized in Table~\ref{tab:notion_notations}. The domain of the hyperparameters $\bp$ depends on the learner $l$.
The test error $\epsilon$ 
is not observable during AutoML. It is 
a blackbox function of the learner $l$, the hyperparameters $\bp$, and the sample size $s$. It is approximated by the validation error $\tilde\epsilon$. 
We observe several non-blackbox relations among the variables, which are not first noticed by us but rarely leveraged by existing AutoML systems.

\begin{table}
\caption{Notions and notations.}
\begin{center} \label{tab:notion_notations}
\small
\begin{tabular}{ |c|c||c|c| }  
 \hline
  $L$ & number of learners & $l$ & learner   \\  \hline
  $\tilde{\epsilon}$ & validation error & $\epsilon$ & test error  \\ \hline
  $\bp$ & hyperparameter values & $\chi$ & configuration \\\hline
  $s$ & sample size & $r$ & resampling strategy   \\\hline
   $M$ & trained model & $\kappa$ & trial cost\\ \hline
\end{tabular}
\end{center}
\end{table}

\begin{observation}[Sample size + resampling \(\rightarrow\) error]\label{ob:sr}
First, it is reasonable to assume the test error $\epsilon$, as well as the gap between $\epsilon$ and $\tilde{\epsilon}$, decreases or stays with the increase of the sample size $s$ when all the other factors are fixed~\citep[pg.~2]{huang2019efficient} (\cite{nakkiran2019deep} observes this after the sample size is above a threshold). Second, the gap between $\epsilon$ and $\tilde{\epsilon}$ is smaller for cross-validation than holdout, when all the other factors are fixed~\cite{Kohavi1995,feurer2015efficient}. 
\end{observation}

\begin{observation}[Hyperparameter + sample size \(\rightarrow\) error]\label{ob:ps}
Many learners have a subset of hyperparameters related to model complexity or regularization, e.g., the number of trees and the depth of each tree
in tree-based learners. 
For a fixed sample size, $\epsilon$ does not necessarily
reach its minimum at maximal complexity. Generally speaking, smaller sample size (in a local region) requires lower complexity and more regularization to avoid overfitting~\cite{hastie01statisticallearning, nakkiran2019deep}.
\end{observation}

\begin{observation}[Quantifiable impact on cost]\label{ob:cost}
For each fixed combination of learner $l$ and resampling strategy $r$, the cost $\kappa$ is approximately proportional to the sample size $s$ and a subset of cost-related hyperparameters, such as the number of trees.
When all the other factors are fixed, k-fold cross-validation roughly takes $\frac{k-1}{1-\rho}\times$ cost as holdout using holdout ratio $\rho$.
\end{observation}

Based on the joint effect of hyperparameter and sample size on error and cost (Observation~\ref{ob:ps} and \ref{ob:cost}), we have the following property. 

\begin{property}[SuitableSampleSize]\label{prop:sample}
Small sample size can be used to train and compare low-complexity configurations, while large sample size is needed for comparing high-complexity configurations. 
\end{property}

From the compound impact of sample size and resampling strategy on error and cost (Observation~\ref{ob:sr} and \ref{ob:cost}), when sample size is small, cross-validation reduces the variance of validation error while the cost is bounded. When sample size is large, validation error from holdout is close to test error, and the cost is much lower than cross-validation. Since the trial-based AutoML requires a fair selection mechanism among all the configurations, we have:

\begin{property}[Resample]\label{prop:resample}
Cross-validation is preferred over holdout for small sample size or large time budget.
\end{property}

From the target of error minimization and Observation~\ref{ob:sr}, as well as the fact that the optimal choice of $l^*$ is  unknown, we derive the following property. 
\begin{property}[FairChance]\label{prop:chance}
Given any search sequence prefix, every learner $l$ should have a chance to be searched again, unless all the valid hyperparameter values of $\bp$ have been searched using the full training data size in the prefix.
\end{property}

From the target of cost minimization and Observation~\ref{ob:cost}, we can derive the following property, which is in general difficult to achieve as the optimal configuration is unknown. 
\begin{property}[OptimalTrial]\label{prop:mincost}
The total cost of any search sequence prefix is desired to be approximately optimal (i.e., have a bounded ratio over the optimal cost) for the lowest error it achieved. Similar for the subsequence corresponding to each learner \(l\). 
\end{property}
Although these properties are idealistic properties and they are not necessarily complete, they provide meaningful guidance in designing a low-cost AutoML system.  

%% file: sec_system.tex
\section{FLAML}\label{sec:system}
We present our design following the guidelines. Section~\ref{sec:overview} presents an overview, and Section~\ref{sec:strategy} details our search strategy used in each component respectively.

\subsection{Design Overview}\label{sec:overview}

\begin{figure}
    \includegraphics[width=1.0\linewidth]{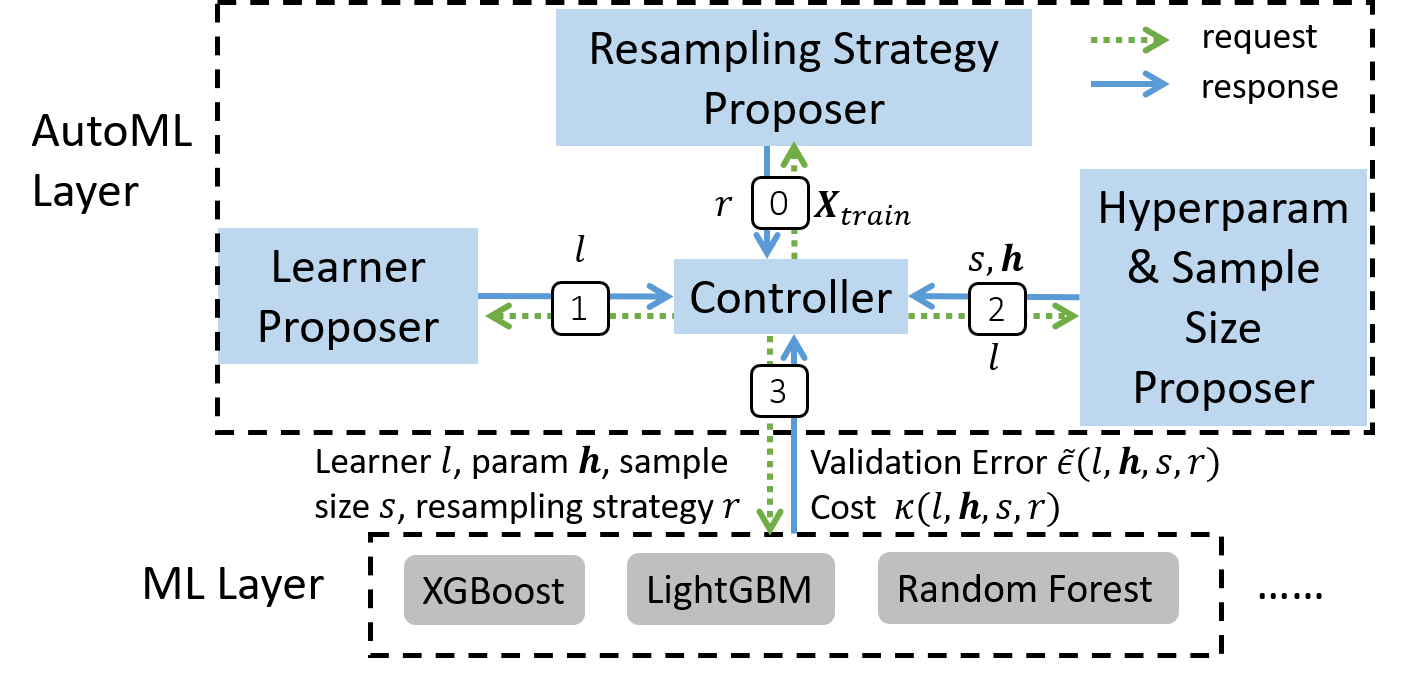}
    \caption{Major components in \FLAML.}
    \label{fig:architect}
\end{figure}

Our design is presented in Figure~\ref{fig:architect}, with the purpose of easy realization of the desired properties described in our analysis. It consists of two layers, including a ML layer and an AutoML layer. The ML layer contains the candidate learners. The AutoML layer includes a learner proposer, a hyperparameter and sample size proposer, a resampling strategy proposer and a controller. The order of the control flow is indexed on the arrows in Figure~\ref{fig:architect} as four steps. Steps 0-2 involve choosing the corresponding variables in each component. 
In step 3, the controller will invoke the trial using the selected learner in the ML layer, and observe the corresponding validation error $\tilde \epsilon$ and cost $\kappa$. Steps 1-3 are repeated by iterations until running out of budget. 
{In a parallel environment, the controller can execute a new iteration of steps 1-3 before an iteration finishes if there are available resources.}
Changing one strategy inside each component does not affect the strategy of others. This design allows easy upgrade by incorporation of novel search schemes to replace each component.

Our system differs from previous work in multiple perspectives: (1) It is different in how we decouple the searched variables and search strategies (Table~\ref{tab:autoML_summary}). For example, we couple the decision of \(\bp\) and \(s\) in our design to ensure sample size is decided together with the hyperparameters, which reflects Property~\ref{prop:sample}. 
We decouple learner and hyperparameters and use the order of \(\{l\}\rightarrow \{\bp,s\}\) to respect domain dependency. 
(2)
As the first trial-based library targeting ad-hoc data (including large-scale datasets) using low-cost, \FLAML\ focuses on the core search efficiency and does not use meta-learning or ensemble. It is considered as future work to develop lightweight meta-learning and ensemble techniques for FLAML while keeping the system economic, generic and capable of handling ad-hoc datasets.  (3) Since the commonly used search strategies are not designed to deeply exploit the compound relations of the multiple factors as analyzed in Section~\ref{sec:analysis}, we employ new search strategies as introduced in the next subsection.

\subsection{Search Strategy}
\label{sec:strategy}

\begin{table*}[t] 
\caption{Comparison of search strategy.} \label{tab:autoML_summary}
\centering\small
\begin{tabular} {r l l}  
  \hline
Tool  & Searched variable & Search strategy \\ 
 \hline
Alpine Meadow~\cite{Shang2019DDS} & $\{l\}\rightarrow\{\bp\}\rightarrow \{s\}$ & Meta learning $\rightarrow$ BO $\rightarrow$ Progressive sampling \\
 Auto-sklearn~\cite{feurer2015efficient} & $\{l,\bp\}$ & Bayesian optimization, with meta-learning and ensemble \\
 H2O AutoML~\cite{H2O} & $\{l\}\rightarrow\{\bp\}$ & Enumeration $\rightarrow$ Randomized grid search, with ensemble \\
HpBandSter~\cite{falkner2018,ICLR:li2017hyperband} & \(\{l,\bp\},\{s\}\) & Bayesian optimization, Hyperband \\
 PMF-automl~\cite{fusi2018probabilistic} & $\{l,\bp\}$ & Collaborative filtering, with post-processing ensemble  \\
 TPOT~\cite{Olson2016TPOT} & $\{l,\bp\}$ & Genetic programming, with ensemble embedded\\
\hline
\FLAML & $\{l\}\rightarrow$ & ECI-based sampling of learner choices \(\rightarrow\) \\
& $\{\bp,s\}$ & Randomized direct search, ECI-based choice of sample size \\
\hline
\end{tabular}
\end{table*}

\begin{figure}
  \centering
       \includegraphics[width=0.98\linewidth]{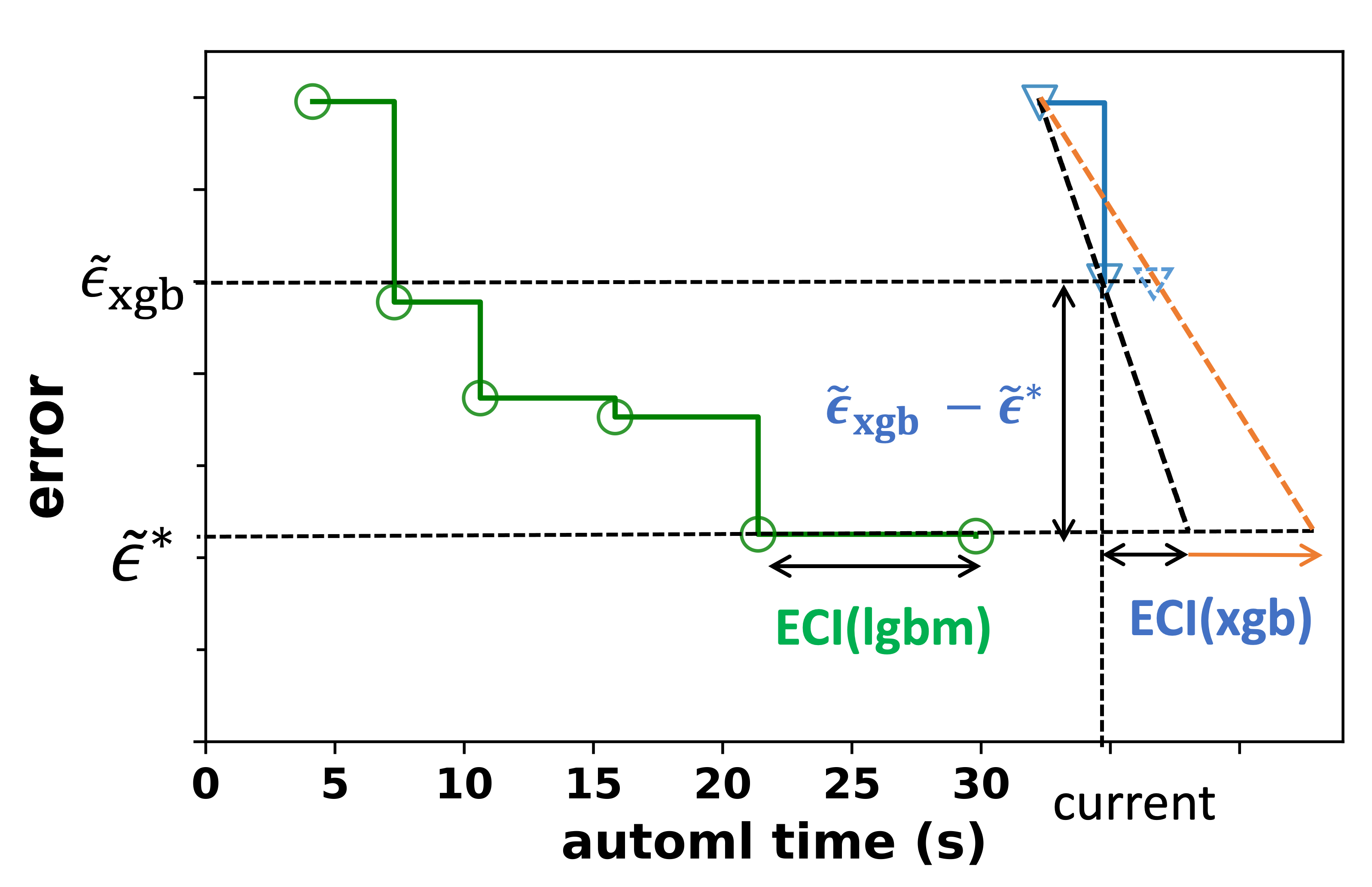} \\
 \hspace*{-0.5cm}\includegraphics[width=1.08\linewidth]{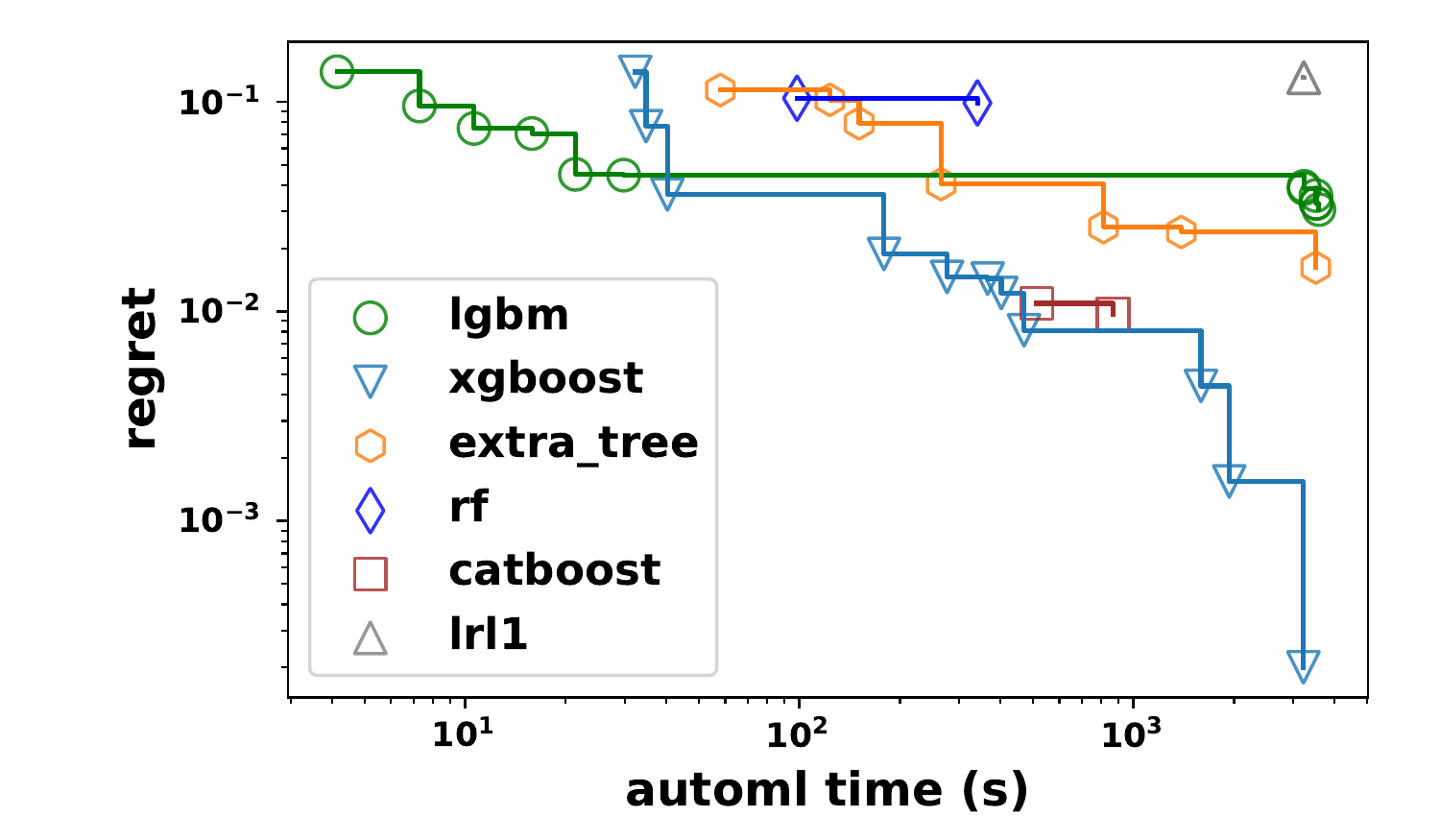}
    \caption{Illustration of ECI-based prioritization. 
    }\label{fig:higgs_eci}
\end{figure}

Before introducing our search strategies, we first introduce the notion of \emph{estimated cost for improvement} (ECI) which will be used in the search strategies. 
For each learner $l\in [L]$, $ECI_1(l)$ (abrv. \(ECI_1\)) denotes our estimation of the cost for $l$ to find a configuration with lower error than the current best error (denoted as $\tilde{\epsilon}_{l}$) under the current sample size. $ECI_2(l)$ (abrv. \(ECI_2\)) denotes our estimation of the cost to try the current configuration for $l$ (which took \(\kappa_{l}\) cost) with increased sample size (multiplied by a factor of \(c\)). Finally, $ECI(l)$ (abrv. \(ECI\)) is our estimation of the cost it takes to find a configuration with \(l\) and lower error than the current best error among all the learners (denoted as $\tilde{\epsilon}^*$). 

Let $K_1>K_2$ be abbreviations of $K_1(l)>K_2(l)$, representing the total cost spent on $l$ when the two most recent updates of best configurations happened for \(l\) respectively, $K_0$ (abbreviations of \(K_0(l)\)) be the total cost spent on $l$ so far, 
and $\delta$ (abbreviations of \(\delta(l)\)) be the error reduction between the two corresponding best configurations. We set:
\begin{equation}
\begin{aligned}
    &ECI_1=\max(K_0-K_1,K_1-K_2), ECI_2=c\cdot\kappa_{l}\\
    &ECI = \max\left(\frac{(\tilde{\epsilon}_{l}-\tilde{\epsilon}^*)(K_0-K_2)}{\delta},\min(ECI_1,ECI_2)\right) 
    \label{eq:ECI}
\end{aligned}
\end{equation}
The calculation of \(ECI_1\) is based on the assumption that it takes higher cost to find an improvement in the later stage of search.\footnote{For learners which have not been tried in the search, the \(ECI_1\) is set to the smallest trial cost for those learners. Since the smallest trial cost varies with input data, we first run the fastest learner and gets its smallest cost on the input data, and then set the \(ECI_1\) for other learners as multiples of this cost using predefined constants. } 
\(ECI_2\) is set to be \(c\) times as large as the trial cost of the current configuration for $l$, because we expect the error of the current configuration to improve when given \(c\) times as large sample size. This simple cost estimation can be refined when the complexity of the training procedure is known with respect to sample size. It works well for the learners in our experiments which have linear complexity.
\(ECI\) is calculated depending on whether $l$ currently has the lowest error among all the learners:  

(a) $l$ currently has the best error among all the learners. In this case, by definition
\(ECI=\min(ECI_1,ECI_2)\).

(b) $l$ does not have the best error among all the learners currently. 
For $l$ to match the current best error $\tilde{\epsilon}^*$, it needs to improve its own error by at least $\tilde{\epsilon}_{l}-\tilde{\epsilon}^*$. 
To estimate the cost to fill that gap, we also need to estimate the \emph{efficiency of improvement} $v$ for $l$. That is, how fast $l$ is expected to reduce the error in its own search sequence. We calculate $v$ as:
\(
    v = \frac{\delta}{\tau}
    \),
where $\tau=K_0-K_2$ is the estimated cost spent on $l$ for making the error reduction \(\delta\). 
In the special case where $\delta=0$, i.e., the first configuration searched for $l$ is the best configuration for $l$ so far, we set $\delta=\tilde{\epsilon}_{l}$, and $\tau$ as the total cost spent on $l$.
In our implementation, we double the cost to fill the gap as the estimated cost for finding an improvement because we assume the improvement with respect to cost has a diminishing return.

Combining the two cases we have Eq.~\eqref{eq:ECI}. A visual demonstration is provided in Figure~\ref{fig:higgs_eci}
    corresponding to the same example in Figure~\ref{fig:casestudy}.
    The figure on the top plots the best error per learner vs. automl time, and visualizes the ECI of two learners LightGBM and XGBoost based on Eq.~\eqref{eq:ECI} when the current time point is 35s. To illustrate that  ECI is self-adjustable, we add a hypothetical new trial of XGBoost (the dashed triangle marker at 38s) which does not find a better model. In this case, ECI(xgb) will be increased as shown in the horizontal orange arrow, and the priority of XGBoost will be decreased. 
    The figure on the bottom visualizes the search trajectory of each learner.

\begin{table*}[h]
    \caption{Details of the case study in Figure~\ref{fig:casestudy}. It reveals that \FLAML avoids trying unnecessarily expensive configs in the beginning more than HpBandSter though they are given the same search space. Even though \FLAML eventually tries expensive configs, it chooses the more promising learner (in this example, XGBoost) based on the observed cost and error in early trials.}
    \label{tab:casestudy}
    \centering\small
    
    \begin{tabular}{r|r|l|c|c|r}
    \hline
    Iter & Time (s) & Learner & Config tried by \FLAML & Error & Cost (s) \\\hline
    1 & 4 & LightGBM & {tree num: 4, leaf num: 4, min child weight: 20, learning rate: 0.1...} & \textbf{0.3272} & 3 \\
    ... & ... & ... & ... & ... & ... \\
    9 & 40 & XGBoost & {tree num: 13, leaf num: 9, min child weight: 18, learning rate: 0.4...} & \textbf{0.2242} & 5 \\
    ... & ... & ... & ... & ... & ... \\
    20 & 402 & XGBoost & {tree num: 76, leaf num: 116, min child weight: 3, learning rate: 0.2...} & \textbf{0.2003} & 26 \\
    ... & ... & ... & ... & ... & ... \\
    26 & 1935 & XGBoost & {tree num: 548, leaf num: 247, min child weight: 1.1, learning rate: 0.02...} & \textbf{0.1896} & 238 \\
    27 & 3225 & XGBoost & {tree num: 1312, leaf num: 739, min child weight: 1.1, learning rate: 0.01...} & \textbf{0.1882} & 1290 \\        
    ... & ... & ... & ... & ... & ... \\

    \hline
    Iter & Time (s) & Learner & Config tried by HpBandSter & Error & Cost (s)\\\hline
    1 & 16 & XGBoost & {tree num: 47, leaf num: 50, min child weight: 0.004, learning rate: 0.8...} & \textbf{0.2497} & 15 \\
    2 & 1193 & XGBoost & {tree num: 17863, leaf num: 2735, min child weight: 3.7, learning rate: 0.1...} & \textbf{0.1979} & 1177 \\
    3 & 1356 & CatBoost & {early stop rounds: 15, learning rate: 0.03...} & \textbf{0.1978} & 163 \\
    ... & ... & ... & ... & ... & ... \\
    7 & 2011 & XGBoost & {tree num: 10369, leaf num: 369, min child weight: 0.1, learning rate: 0.4...} & 0.2036 & 583 \\
    8 & 3325 & RF & {tree num: 2155, max features: 0.36, criterion: entropy} & {0.2007} & 1313 \\
    ... & ... & ... & ... & ... & ... \\ \hline
    \end{tabular}    
\end{table*}

\textbf{Step 0: The resampling strategy proposer chooses $r$.} 
Resampling strategy is decided based on a simple thresholding rule. It is the simplest design which follows Property~\ref{prop:resample} (Resample). If the training dataset has fewer than 100K instances and \# instances \(\times\) \# features \(/\) budget is smaller than 10M/hour, we use cross validation. Otherwise, we use holdout. This simple thresholding rule can be easily replaced by more complex rules, e.g., from meta learning. By default, \FLAML\ uses 5-fold cross-validation and 0.1 as the holdout ratio.

\textbf{Step 1: The learner proposer chooses $l$.} 
With the concept of ECI introduced, we design a search strategy where each learner $l$ is chosen with probability proportional to $1/ECI(l)$.
There are several reasons why ECI is desirable in our system: (1) This design follows Property~\ref{prop:chance} and \ref{prop:mincost}. Property~\ref{prop:mincost} (OptimalTrial) suggests that we prioritize choices which are expected to improve the error using small cost, hence we assign choices with lower ECI higher probability. (2) Instead of directly choosing the learner with lowest ECI, we use randomization because 
Property~\ref{prop:chance} (FairChance) requires every learner to have chance to be searched again, and our estimation is not precise. Based on our sampling scheme, the expectation of ECI for the probabilistic choice is 
$E[ECI] = \sum_{l} \frac{ECI(l)\cdot ECI(l)^{-1}}{\sum_{l'} ECI(l')^{-1}} =$
the harmonic mean of all the ECIs. That means, the expected cost for improvement using our sampling scheme is still dominated by and close to the lowest ECIs.
(3) With more observations about $l$ being collected, ECI will be updated dynamically. The dynamic update of ECI leads to a self-correcting behavior: If our ECI is an overestimate, it will decrease; if it is an underestimate, it will increase. 
This can be reflected from the formula of ECI and Figure~\ref{fig:higgs_eci}. 

Although a related concept EIperSec (Expected Improvement per Second) was proposed in~\cite{snoek2012practical}, it is designed for 
a different context of Bayesian optimization 
and not applicable to our goal of learner selection. 

\textbf{Step 2: The hyperparameter and sample size proposer chooses $\bp$ and $s$.}

For hyperparameters, we adopt a recently proposed randomized direct search method 
~\cite{wu2021cost}, which can perform cost-effective optimization for cost-related hyperparameters. 
The algorithm uses a low-cost configuration as the start point. At each iteration, it samples a random direction $\bu$ in a $(|\bp|-1)$-dimensional unit sphere and then decides whether to move to a new $\bp$ along the randomly sampled direction (or the opposite direction) depending on the observed sign of change of validation error. The cost of the next trial can be upper bounded with respect to the cost of the best config of the considered learner. This upper bound of trial cost is guaranteed by the search procedure used, and increases only progressively if the best error is reduced. Step-size of the move is adjusted adaptively (large in the beginning to fast approach the required complexity) and the search is restarted (from randomized initial points) occasionally to escape local optima.

Though the randomized direct search method does not handle subsampling,
it is a good option to use in our framework due to two important reasons: (1) The method proved its effectiveness in controlling trial cost both theoretically and empirically. The theoretical analysis of this method shows its alignment with Property 4 (OptimalTrial), and its empirical study demonstrates superiority over Bayesian optimization methods including the one using EIPerSec when cost-related hyperparameters exist. (2) The method works without requiring the exact validation error of each trial as feedback, as long as the relative order of any two trials can be determined, and we can leverage that to modify the method to incorporate data subsampling.
We make several important adjustments to enable data subsampling. Specifically, we begin with a small sample size (10K) for each \(l\). For each requested \(l\), we first make a choice between increasing the sample size and trying a new configuration with the current sample size, by comparing $ECI_1(l)$ and $ECI_2(l)$. 
When $ECI_1(l)\ge ECI_2(l)$, we keep the current hyperparameter values and increase the sample size. Otherwise, we stay with the current sample size, and generate hyperparameter values using the randomized direct search method described above. With this design, the system will adaptively change the sample size as needed. Once the sample size for a learner reaches the full data size, it keeps using that size until convergence for that learner. That reduces the risk of pruning good configurations by small sample size compared to multi-fidelity pruning. We reset the sample size to the initial value as the search for that learner is restarted. 

 The implementation of the randomized direct search method follows~\cite{wu2021cost}. 
 At each iteration, a random direction is used first to train a model. If the error does not reduce, we train another model using the opposite direction. The initial stepsize is set to be $\sqrt{d}$. It will be decreased when the number of consecutively no improvement iterations is larger than $2^{d-1}$ until it reaches a lower bound, i.e., converges. Specifically, stepsize is discounted by a reduction ratio \(>1\), which is intuitively the ratio between the total number of iterations taken in total since the last restart of the search and the total number of iterations taken to find the best configuration since the last restart. 
 We perform adaptive step-size adjustments and random restart only when the largest sample size is reached. 
 Our system shuffles the data randomly in the beginning and to get a sample with size $s$, it takes the first $s$ tuples of the shuffled data. Stratified shuffling is used for classification tasks based on the labels.

\textbf{Advantages of our design}. First, our search strategies are designed toward strong \emph{final performance} (i.e., low final error) for ad-hoc datasets, which requires a large configuration search space. The random sampling according to ECI in Step 1 and the random restart in Step 2 help the method escape local optima. 
Second, our search strategies are designed toward strong \emph{anytime performance} for ad-hoc datasets. 
The ECI-based prioritization in Step 1 favors cheap learners in the beginning but penalizes them later if the error improvement is slow. The hyperparameter and sample size proposer in Step 2 tends to 
propose cheap configurations at the beginning stage of the search, but quickly move to configurations with high model complexity and large sample size when needed in the later stage of the search. 
These designs make FLAML navigate large search space efficiently for both small and large datasets.
Last, the computational overhead in the AutoML layer compared to the trial cost in the ML layer is negligible in our solution: ECI-based sampling, randomized direct search, and update of ECIs. The complexity of these operations for each iteration is linear with the dimensionality of hyperparameters,
and does not depend on the number of trials.

%% file: sec_exp.tex
\section{Experiments}

\begin{figure*}[t] 
  \centering

    \begin{subfigure}{\linewidth}
    \centering
    \includegraphics[width=\linewidth]{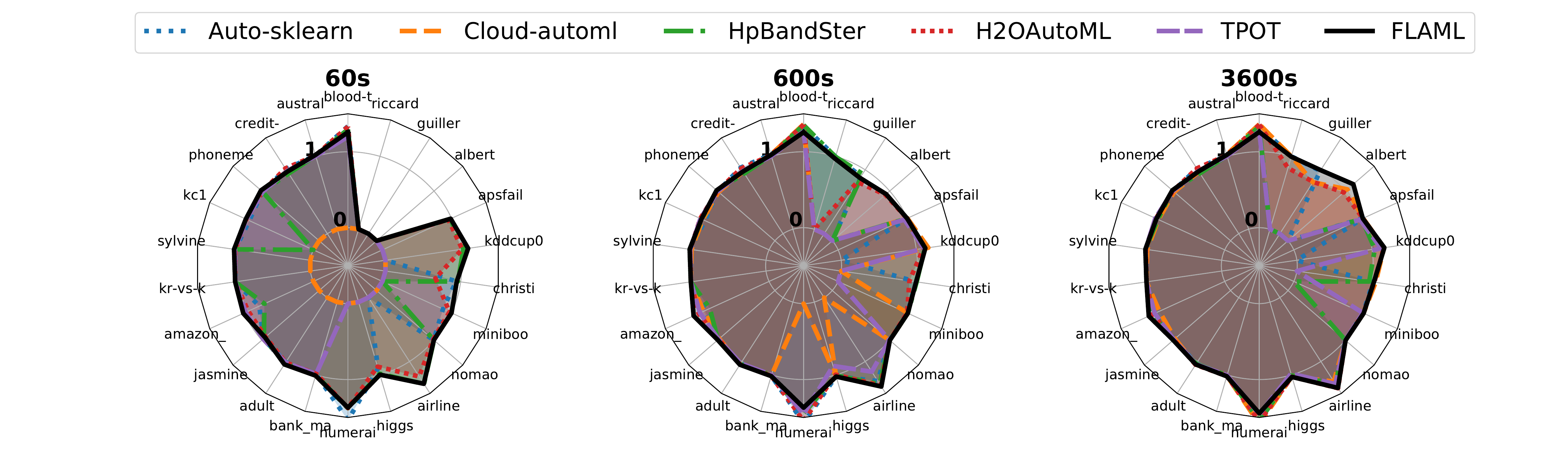}
    \caption{Binary classification datasets ordered by size counter clockwise, from smallest \textit{blood-transfusion} to largest \textit{riccardo} }
  \end{subfigure} 
  
  \begin{subfigure}{\linewidth}
    \centering
    \includegraphics[width=\linewidth]{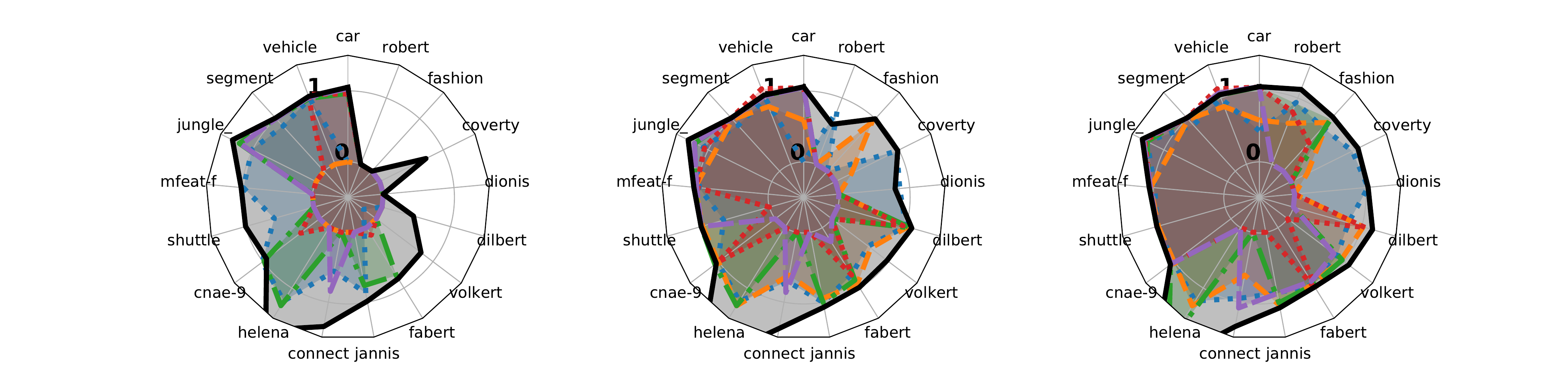}
    \caption{Multi-class classification datasets ordered by size counter clockwise, from smallest \textit{car} to largest \textit{robert}}
  \end{subfigure}

   \begin{subfigure}{\linewidth}
    \centering
    \includegraphics[width=\linewidth]{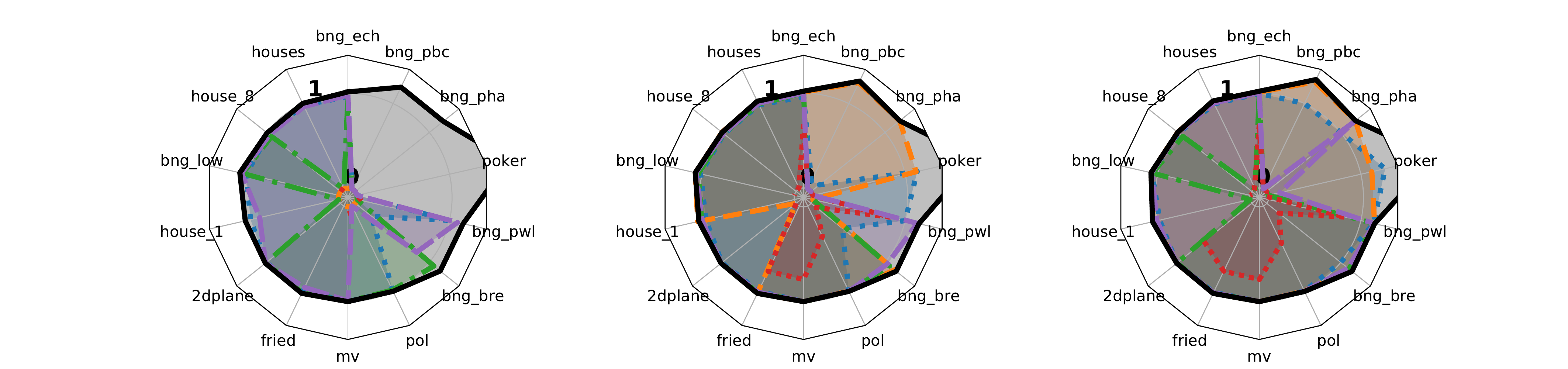}
    \caption{Regression datasets ordered by size counter clockwise, from smallest \textit{bng\_echomonths} to largest \textit{bng\_pbc}}
  \end{subfigure} 
    
  \caption{Scaled scores of AutoML libraries on each dataset with each time budget. The longer is each spoke the better.}  
  \label{fig:FLAML_radar_chart}
\end{figure*} 

\begin{figure*}[t]
  \centering
  \begin{subfigure}{0.39\textwidth}
\includegraphics[width=\columnwidth]{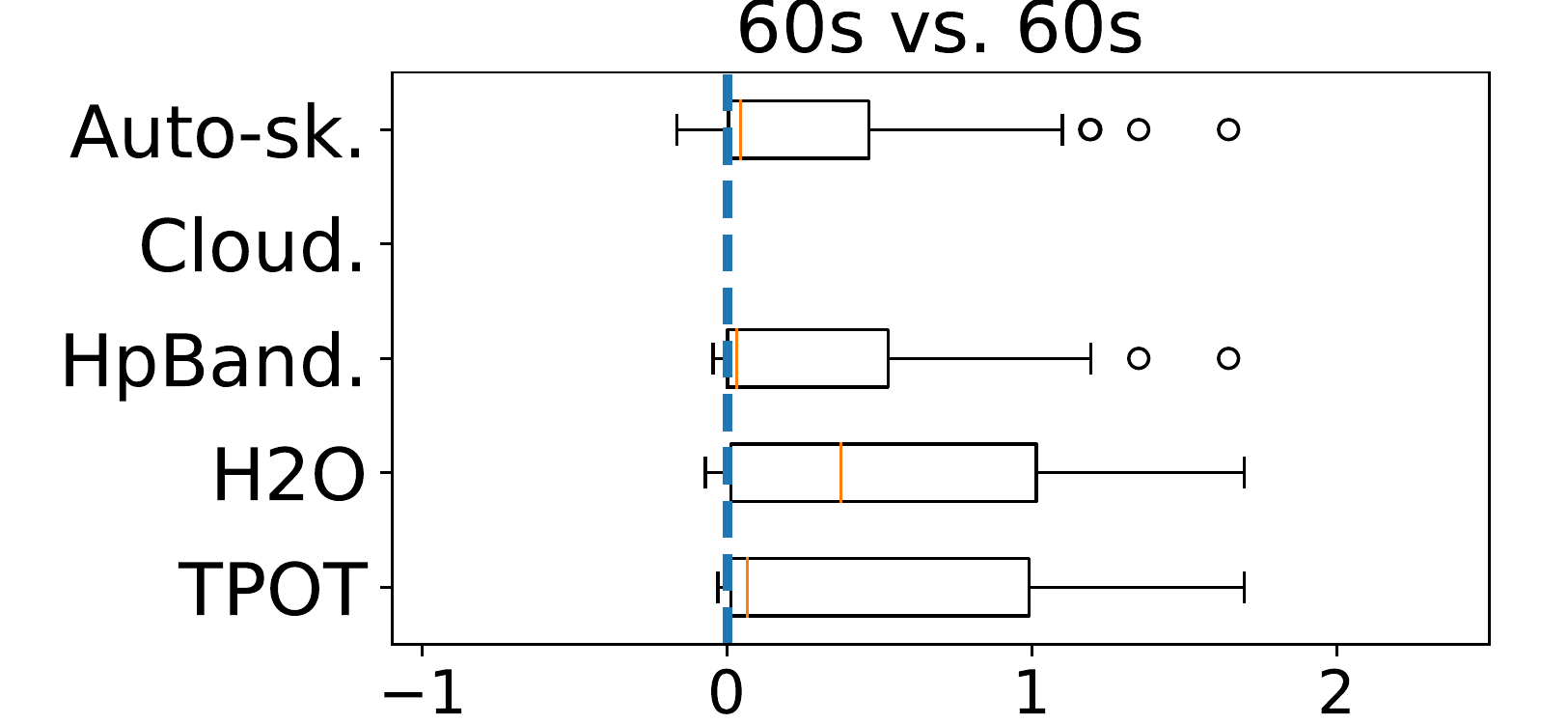}%
\end{subfigure}\hfill%
\begin{subfigure}{0.30\textwidth}
\includegraphics[width=\columnwidth]{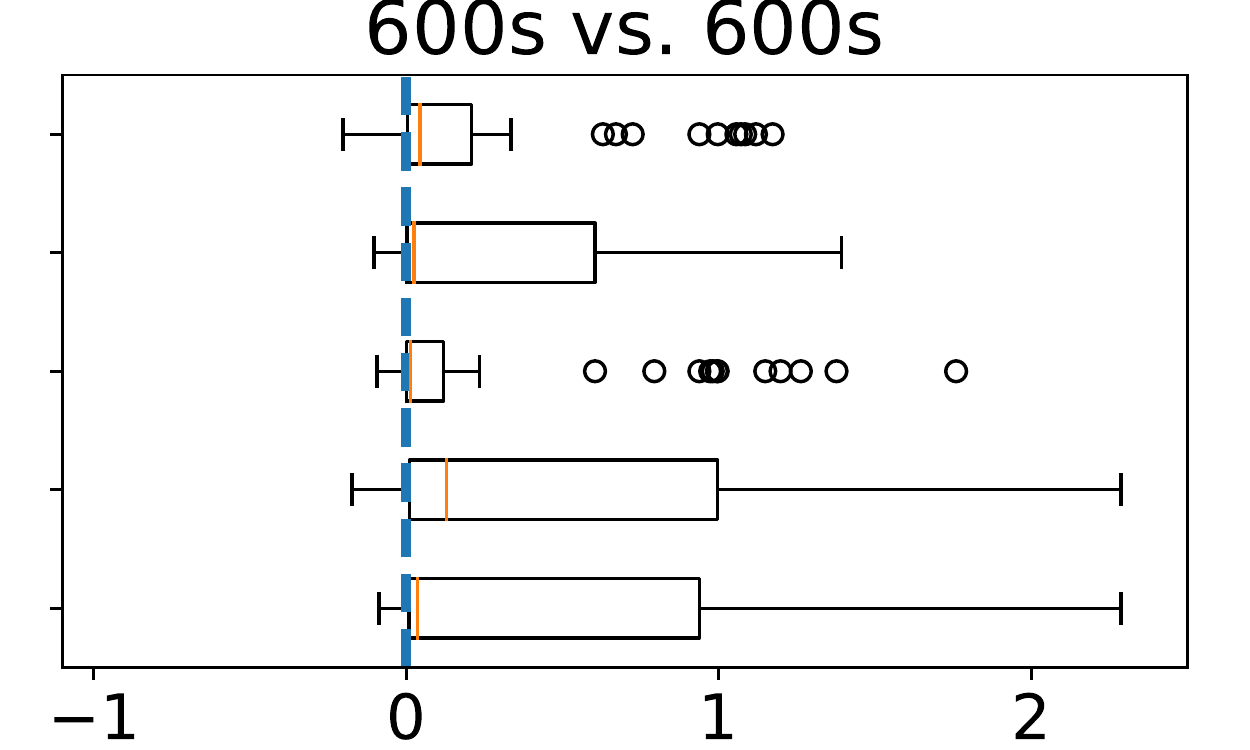}%
\end{subfigure}\hfill%
\begin{subfigure}{0.30\textwidth}
\includegraphics[width=\columnwidth]{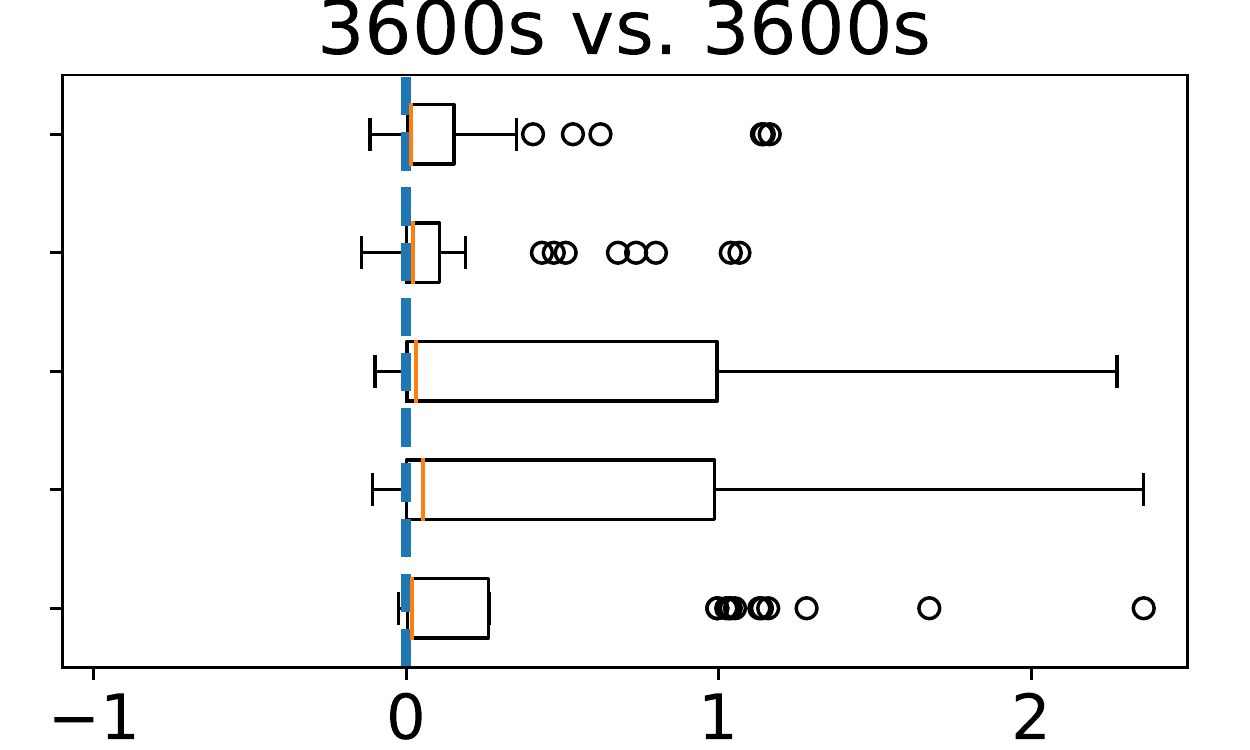}%
\end{subfigure}

  \begin{subfigure}{0.39\textwidth}
\includegraphics[width=\columnwidth]{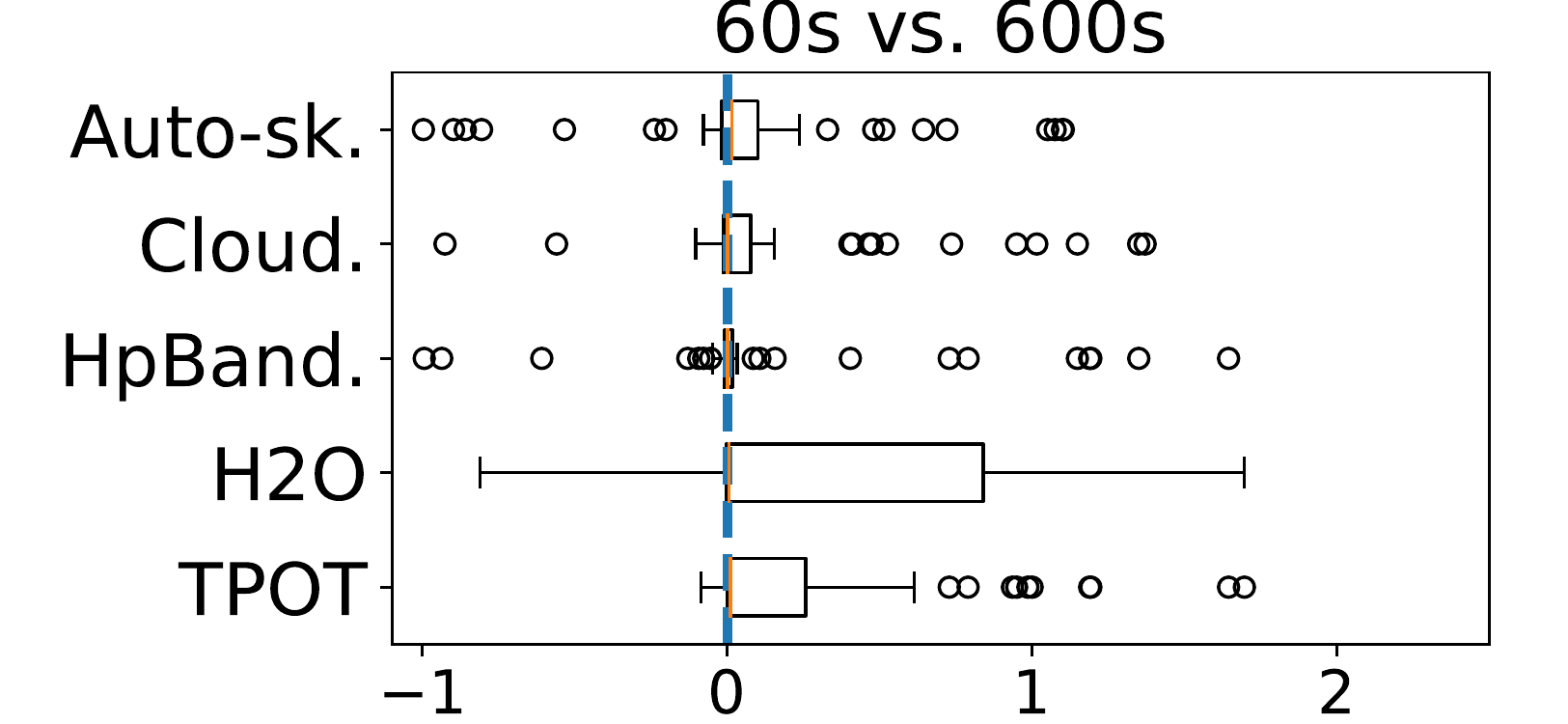}%
\end{subfigure}\hfill%
\begin{subfigure}{0.30\textwidth}
\includegraphics[width=\columnwidth]{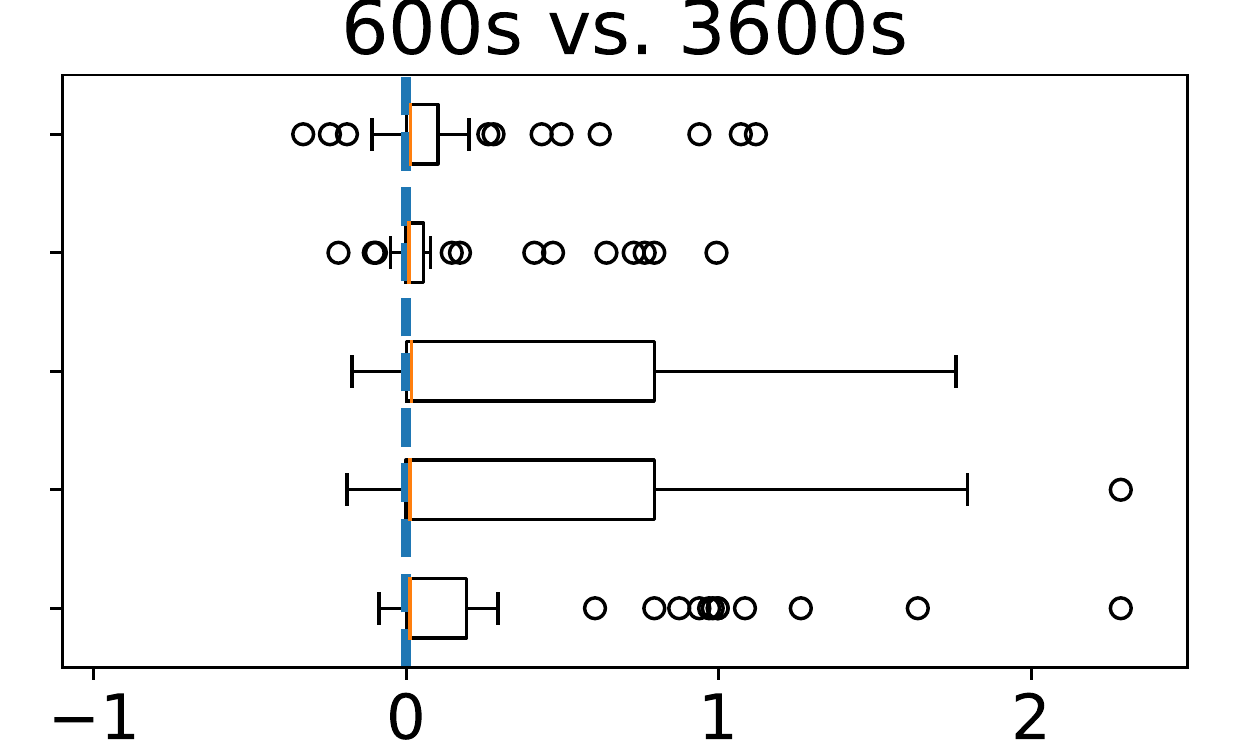}%
\end{subfigure}\hfill%
\begin{subfigure}{0.30\textwidth}
\includegraphics[width=\columnwidth]{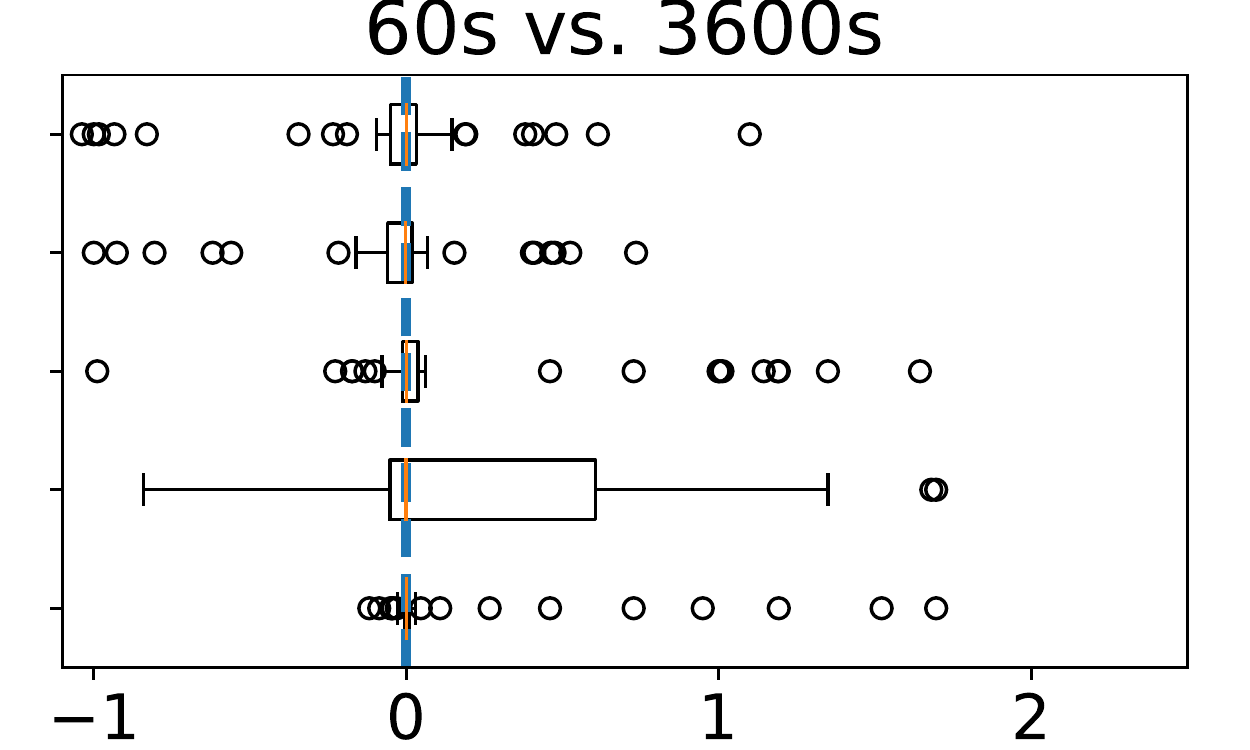}%
\end{subfigure}\hfill%
  \caption{Box plot of scaled score difference between \FLAML\ and other libraries when \FLAML\ uses equal or smaller budget (positive difference meaning \FLAML\ is better).}
  \label{fig:margin}
\end{figure*}

Our main empirical study is based on a combination of a recent open source AutoML classification benchmark~\cite{Gijsbers2019benchmark} and a regression benchmark~\cite{Olson2017PMLB}, for a total of 53 datasets (39 classification + 14 regression). 
{
The two benchmarks can be found at: 
AutoML Benchmark (classification) -
\url{https://openml.github.io/automlbenchmark}, and
PMLB (regression) - 
\url{https://github.com/EpistasisLab/penn-ml-benchmarks}.
}
The first benchmark collects 39 classification tasks that represent real-world data science problems of various sizes, domains and levels of difficulty from previous AutoML papers, competitions and benchmarks. 
In the second benchmark PMLB, most datasets are of small scale,
from which we selected the regression datasets whose numbers of instances are larger than 10,000. That results in 14 regression tasks.
The statistics of all the 53 datasets are listed in Table~\ref{tab:data}-\ref{tab:regdata} in the appendix. 
The 53 datasets have \#instance $\times$ \#feature ranging from 2,992 to 85,920,000 and vary in the occurrence of numeric features, categorical features and missing values. 
The referred AutoML classification benchmark uses roc-auc and negative log-loss as the scores to evaluate the performance on binary classification tasks and multi-class classification tasks respectively.{ It calibrates the original scores using a constant class prior predictor (=0) and a tuned random forest (=1), the higher the better. The tuned random forest is a strong baseline taking a long time to finish, and achieving a score above 1 is not easy according to \cite{Gijsbers2019benchmark}.
For regression tasks, 
we use the r2 score
which is a metric bounded by 1 before calibration. }

We compare FLAML (v0.1.3) to four trial-based AutoML libraries plus a hyperparameter optimization library designed for budget constrained scenarios: auto-sklearn (v0.9.0)\footnote{AutoSklearn2Classifier~\cite{ASKL2} for classification.}, H2O AutoML (v3.30.0.3), TPOT (v0.10.1), cloud-automl (a commercial cloud-based AutoML service from one major cloud provider), and
HpBandSter (v0.7.4, an implementation of BOHB which integrates Bayesian optimization with Hyperband). The first three are the top three performers reported by the AutoML benchmark~\cite{Gijsbers2019benchmark} and their performance is close to each other. 
HpBandSter uses the same search space and resampling strategy as those of \FLAML (Table~\ref{tab:searchspace} in the appendix).{ It is worth noting that all the libraries use a different search space from each other by design except for HpBandSter and \FLAML. The search space of \FLAML (reported in the appendix) neither subsumes, nor is subsumed by the search space of auto-sklearn, cloud-automl, H2O AutoML or TPOT. It is very challenging, if not impossible, to equalize the search space due to the specific designs of each library (e.g., meta-learning by auto-sklearn and cloud-automl, and special grid search by H2O AutoML). 
We use their default setting and do not introduce our own bias.}

All experiments are executed on an Ubuntu server with Intel Xeon Gold 6140 2.3GHz, 
and 512GB RAM. We use 1 CPU core for each compared solution and vary the time budget from one  minute to one hour.{ We choose these settings because this work is mainly concerned about performance using low resource, while the numbers in~\cite{Gijsbers2019benchmark} are obtained using a rather generous budget (8 to 32 hours of total CPU time). 
Cloud-automl with 1m budget is not reported since it does not return within 2 minutes.}
As each dataset has been split into 10 folds by OpenML~\cite{Vanschoren2014}, all the reported results are averaged over the 10 folds.

\begin{figure*}
  \centering
  \begin{subfigure}{0.33\textwidth}
\includegraphics[width=\columnwidth]{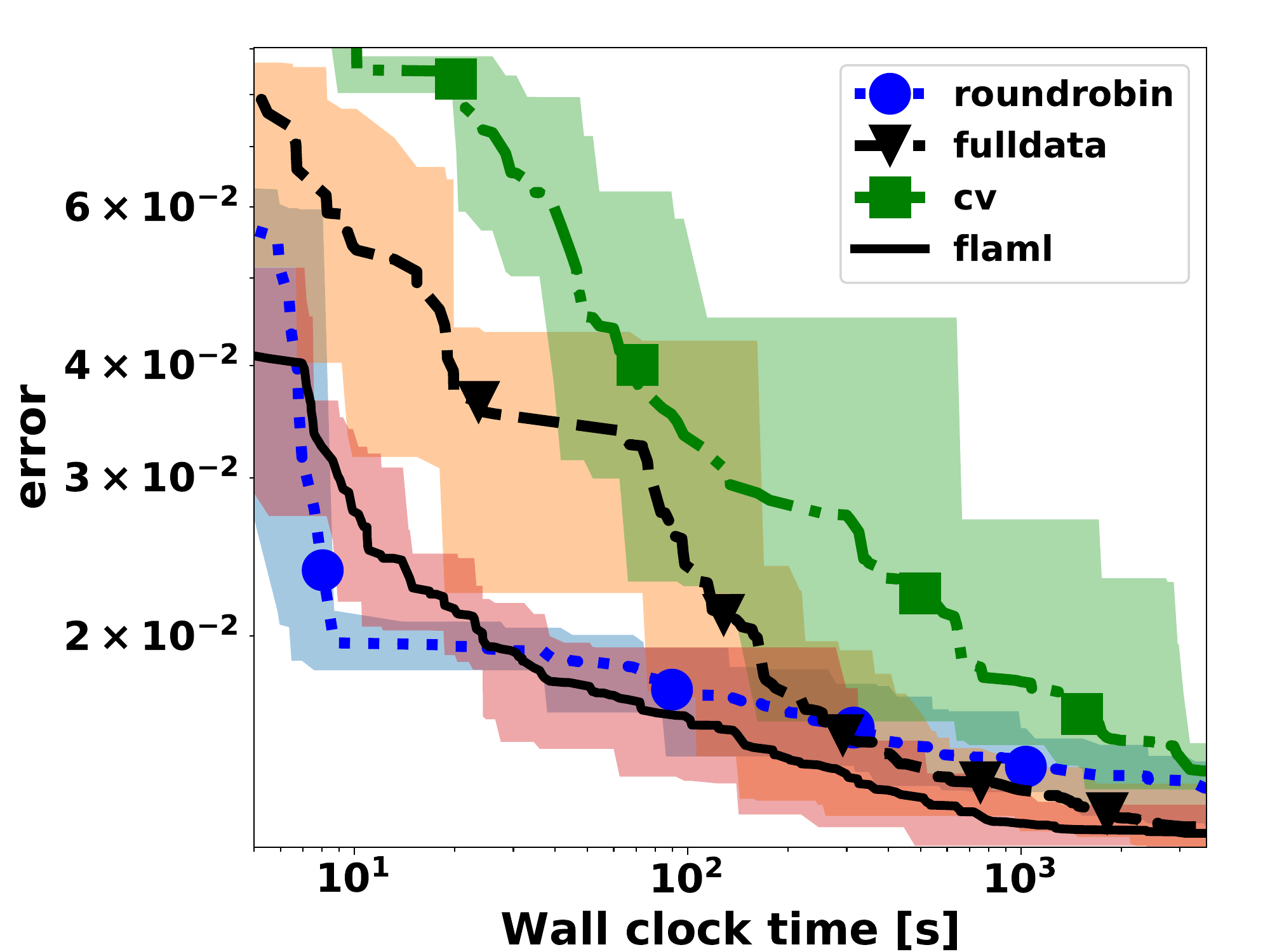}%
\caption{MiniBooNE: 1-auc}
\end{subfigure}\hfill%
\begin{subfigure}{0.33\textwidth}
\includegraphics[width=\columnwidth]{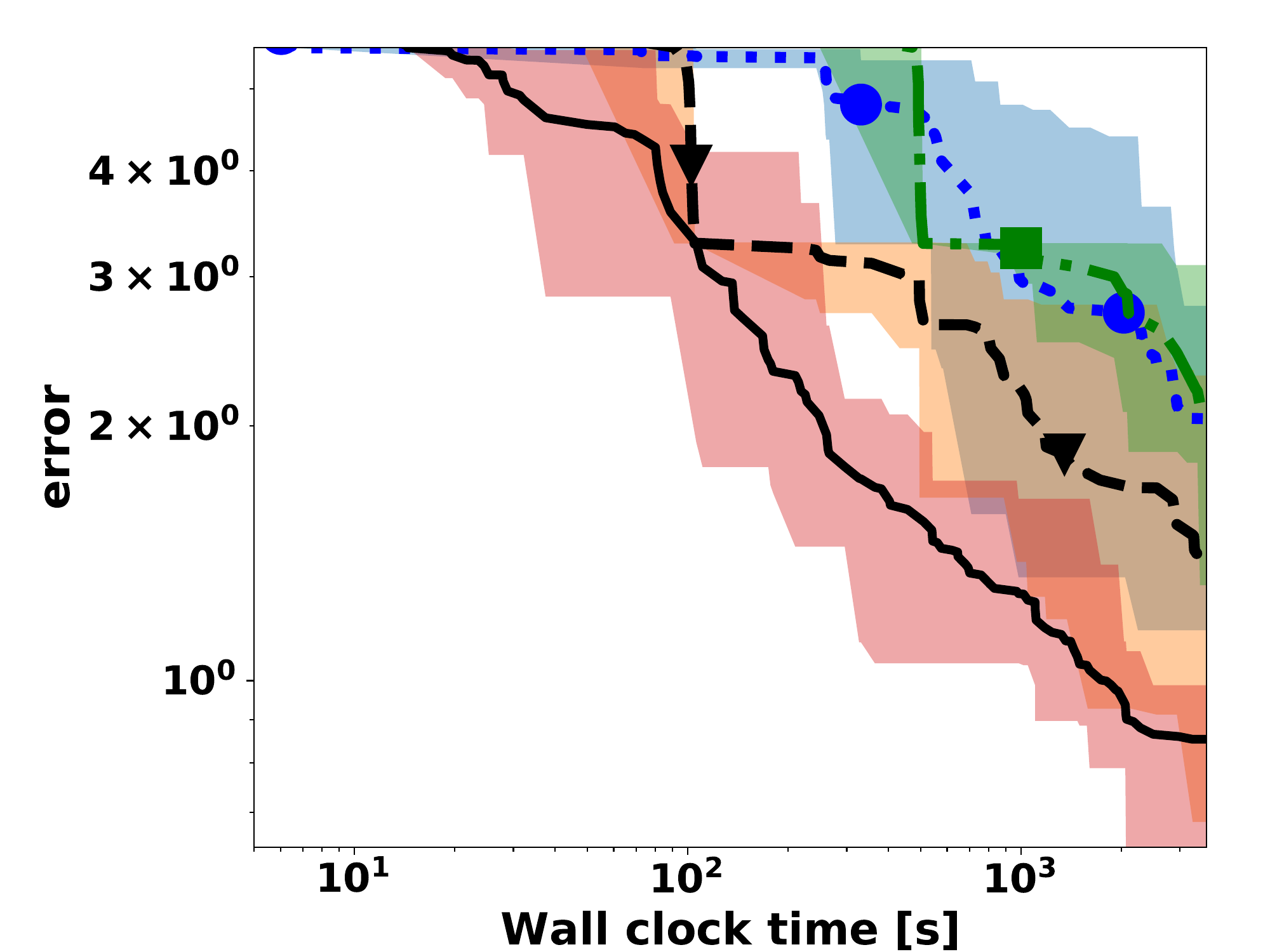}%
\caption{Dionis: logloss}
\end{subfigure}\hfill
\begin{subfigure}{0.33\textwidth}
\includegraphics[width=\columnwidth]{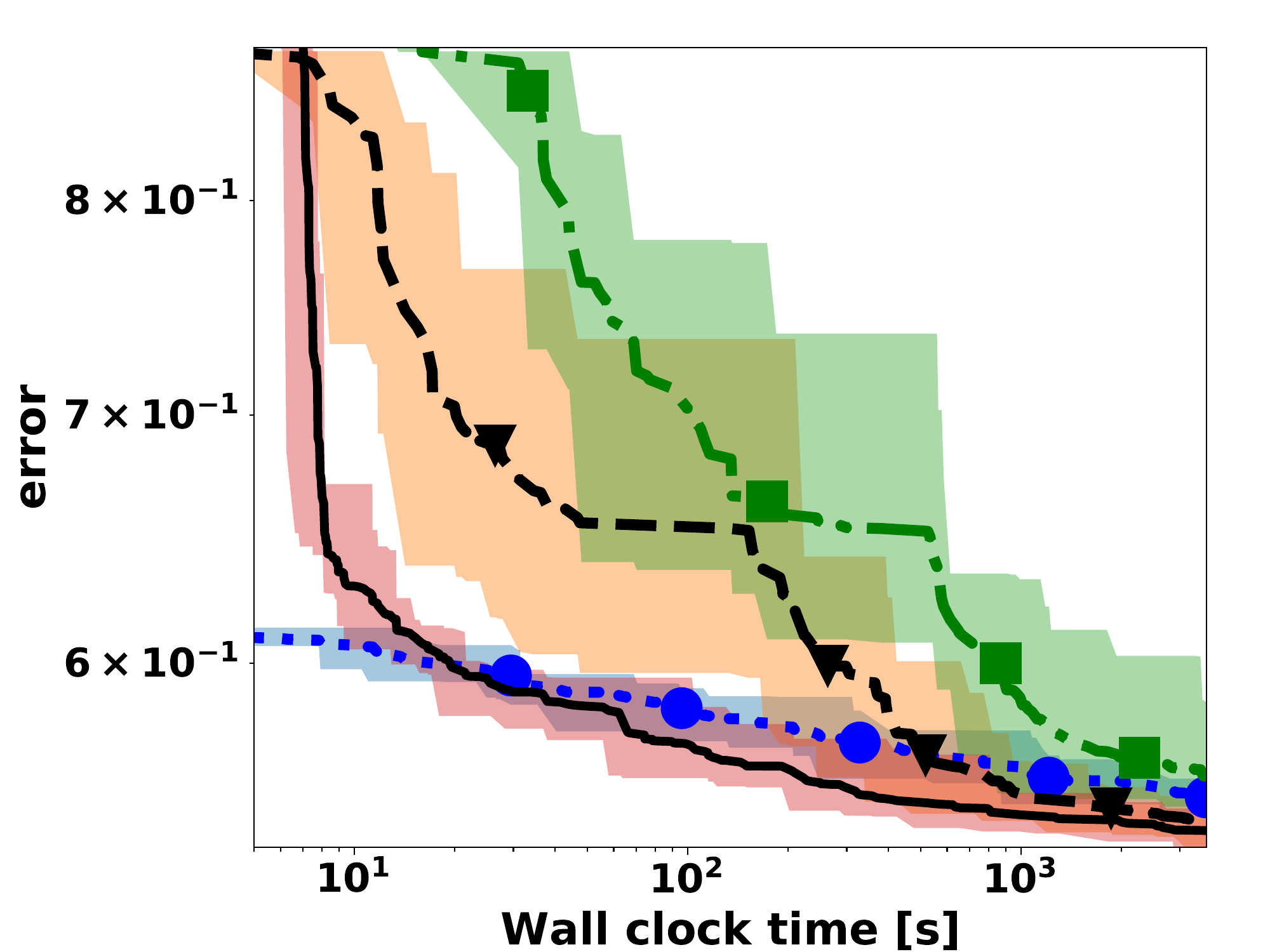}%
\caption{bng\_pbc: 1-r2}
\end{subfigure}\hfill%
\caption{Variations of \FLAML. Lines -- average error over 10 folds; shades -- max and min.}
  \label{fig:curve}
\end{figure*}

\subsection{Comparative Study}
The scaled scores of all the methods given different desired time budgets (1m, 10m and 1h) on all the datasets are shown in Figure~\ref{fig:FLAML_radar_chart}. Each of the radar charts shows the scaled scores of  different methods on a group of datasets (spokes on the radar chart) given a desired time budget. Results of all 53 datasets are summarized into 3 sub-figures (rows) according to their task types. Each row shows the performance comparison on the same group of datasets given different desired time budgets. 
Figure~\ref{fig:margin} presents the distribution of score difference between \FLAML\ and each individual baseline, under equal budget (the first row) or smaller budget for \FLAML\ (the second row). 

When using equal time budgets, \FLAML\ clearly outperforms every competitor with large margins in most cases. {
In a small fraction of cases, \FLAML\ underperforms by a small margin.
}
Even with a smaller time budget, \FLAML\ can be better than or equal to the others in many cases. For example, \FLAML's 1m result is no worse than others' 10m result on 62\%-83\% datasets, and 72\%-89\% for 10m vs. 1h. 
In sum, \FLAML\ demonstrates
\emph{significant margin over each competitor given equal budgets, and
competitive performance given smaller budgets than competitors}. 
To be fair, the prior libraries are not primarily designed for the same low-resource setting as ours.
We note that the search space of \FLAML contains both cheap and expensive configurations, but our integrated search strategies make \FLAML wisely prioritize them. 
As seen in Figure~\ref{fig:casestudy} and Table~\ref{tab:casestudy}, \FLAML's adaptive behavior is the key to strong anytime performance. 
%

\subsection{Ablation Study}

We study the effect of the three components in our system, 
by comparing \FLAML\ with three alternatives: \textit{roundrobin}, which takes trials for each learner $l$ in turn; \textit{fulldata}, which uses the full data for the trials; and \textit{cv}, which uses cross validation for all the trials.
Figure~\ref{fig:curve} plots the validation error of \FLAML\ and its alternatives on a binary classification dataset \emph{MiniBooNE}, a multi-class classification dataset \emph{Dionis}, and a regression dataset \emph{bng\_pbc}. The figure shows how the error improves with respect to search time. We can see that when removing any of the three aspects of the \FLAML\ search strategy, the search performance degrades. In particular, the gap between \FLAML\ and \textit{roundrobin} increases before converging
due to 
\FLAML's prioritization to more promising learners. The gap between \FLAML\ and \textit{fulldata} is large initially because the initial trials of \FLAML\ are very fast using small sample of training data. That gap reduces later as \FLAML\ increases its sample size. 

The ablation study verifies the effectiveness of the search strategies.  Figure~\ref{fig:ablation_box} in the appendix plots the score difference between \FLAML\ and these alternatives over all the datasets.

\subsection{Application to Selectivity Estimation}
\label{sec:ce}

As an example application to database systems, we evaluate the performance of AutoML libraries to the selectivity estimation task. Selectivity estimates are necessary inputs for a query optimizer
to identify a good execution plan~\cite{Selinger79}. A
good selectivity estimator should provide accurate and fast
estimates for a wide variety of intermediate query expressions
at reasonable construction overhead~\cite{Cormode12}. Estimators
in most database systems make use of limited statistics on the data, e.g., per-attribute histograms or small data samples
on the base tables~\cite{oracle_stats,sqlserver_stats,Neumann20umbra}.
\cite{Dutt2019selectivity, dutt2020efficiently} developed low-overhead regression models which achieve much lower q-error (a relative error metric used by the selectivity estimation literature) than all the other methods with similar inference time, including the ones used in commercial database products. The recommended learner is XGBoost with 16 trees and 16 leaves per tree. We denote this configuration as `Manual'. 

\begin{table}
\caption{95th-percentile q-error for selectivity estimation (search time listed if at least one method exceeds time limit). H2O AutoML cannot return a model with the given budget.}
\centering\small
\begin{tabular} {r| c | c| c|c } 
Dataset  &  \FLAML & Auto-sk.  & TPOT & Manual\\ 
 \hline
 2D-Forest  & \textbf{1.41} &   1.42    &   2.70 & 1.84   \\
 2D-Power   & \textbf{2.03} &   3.28    &   4.70 & 4.09    \\
 2D-TPCH    & 2.19(42s) &   \textbf{2.11}(197s)    &   N/A & 3.04\\
 4D-Forest1   &   \textbf{2.91}   &   4.33    &   11.9  & 4.41    \\
 4D-Forest2   &   \textbf{4.40}(45s)   &   5.93(55s)    &   17.4(146s)  & 6.26  \\
 4D-Power   &   \textbf{2.42}(49s)   &   3.78(197s) &   12.4(79s)    & 4.29    \\
 7D-Higgs   &   \textbf{3.16}(60s)   &   5.83(55s)    &   9.65(91s)    & 6.54    \\
 7D-Power   &   \textbf{4.25}(46s)   &   6.87(55s)    &   65.2(102s)    & 7.57    \\
 7D-Weather &   \textbf{4.71}(54s)   &   6.44(55s)    &   30.1(118s)    & 6.84    \\
 10D-Forest & \textbf{9.09}(49s)   &   19.8(147s)  &   96.2(89s)    &   15.1    \\
\end{tabular}
\label{tab:qerror}
\end{table}

Table~\ref{tab:qerror} compares the 95th-percentile q-error of the models found by different AutoML libraries with one CPU minute budget, using the same datasets from~\cite{Dutt2019selectivity}.
\FLAML\ outperforms the other AutoML libraries as well as the manual configuration. 
On 10D-Forest, \FLAML\ is the only AutoML solution that outperforms Manual.

%% file: sec_conclusion.tex
\section{Future Work}\label{sec:future}
While \FLAML\ has superior performance on a variety of benchmark tasks compared to the state of the
art, 
it does not use meta learning to optimize per task instance based on previous experience. It is worthwhile to think how to leverage meta learning in the cost-optimizing framework without losing the robustness on ad-hoc datasets. 
Similarly, it is interesting to study the new tradeoff between cost and error when model ensembles are introduced.

%% file: sec_appendix.tex
\newpage
\begin{table}
\caption{Default search space in \FLAML. \(S\) denotes the number of training instances. Bold values indicate initialization corresponding to lowest complexity and cost. lr - logistic regression.} \label{tab:searchspace}
\centering
\small
\begin{tabular} {c | c| c |c } 
Learner & Hyperparameter  & Type & Range \\ [0.5ex]
 \hline
 & tree num & int & [\textbf{4}, min(32768,S)]\\ 
& leaf num & int & [\textbf{4}, min(32768,S)]\\ 
& min child weight & float & [0.01, \textbf{20}]\\
& learning rate & float & [0.01, 1.0] \\
XGBoost & subsample & float & [0.6, 1.0]  \\
& reg alpha & float & [1e-10, 1.0]   \\
& reg lambda & float & [1e-10, 1.0] \\
& colsample by level & float & [0.6, 1.0]  \\
& colsample by tree & float & [0.7, 1.0]  \\
 \hline
 & tree num & int & [\textbf{4}, min(32768,S)]\\ 
& leaf num & int & [\textbf{4}, min(32768,S)] \\ 
& min child weight & float & [0.01, \textbf{20}] \\
& learning rate & float & [0.01, 1.0] \\
LightGBM & subsample & float & [0.6, 1.0]  \\
& reg alpha & float & [1e-10, 1.0]   \\
& reg lambda & float & [1e-10, 1.0] \\
& max bin & float & [7, 1023]  \\
& colsample by tree & float & [0.7, 1.0] \\
\hline
CatBoost & early stop rounds & int & [\textbf{10}, 150] \\
        & learning rate & float & [0.005,0.2] \\
\hline 
sklearn & tree num & int & [\textbf{4}, min(2048,S)]\\ 
random & max features  & float & [0.1, 1.0]\\ 
forest  & split criterion & cat & \{gini, entropy\}\\
\hline
sklearn & tree num & int & [\textbf{4}, min(2048,S)]\\ 
extra & max features  & float & [0.1, 1.0]\\ 
trees  & split criterion & cat & \{gini, entropy\}\\
\hline
sklearn lr & C & float & [0.03125, 32768]\\
\end{tabular}
\end{table}

\begin{table}
\caption{Binary classification datasets.} \label{tab:data}
\centering
\small
\begin{tabular} {  l| l |r| l } 
name & task id & \# instance  & \# feature \\
\hline
adult  & {7592} & 48842 & 14 \\
Airlines & {189354} & 539383 & 7 \\
Albert & {189356} & 425240 & 78 \\
Amazon\_employee\_access & {34539} & 32769 & 9 \\
APSFailure & {168868} & 76000 & 170 \\
Australian & {146818} & 690 & 14 \\
bank\_marketing & {14965} & 45211 & 16 \\
blood-transfusion & {10101} & 748 & 4 \\
christine & {168908} & 5418 & 1636 \\
credit-g & {31} & 1000 & 20 \\
guillermo & {168337} & 20000 & 4296 \\
higgs & {146606} & 98050 & 28 \\
jasmine & {168911} & 2984 & 144 \\
kc1 & {3917} & 2109 & 21 \\
KDDCup09\_appetency & {3945} & 50000 & 230 \\
kr-vs-kp & {3} & 3196 & 36 \\
MiniBooNE & {168335} & 130064 & 50 \\
nomao & {9977} & 34465 & 118 \\
numerai28.6 & {167120} & 96320 & 21 \\
phoneme & {9952} & 5404 & 5 \\
riccardo & {168333} & 20000 & 4296 \\
sylvine & {168853} & 5124 & 20 \\
\end{tabular}
\end{table}
\begin{table}
\caption{Multi class classification datasets.} \label{tab:multidata}
\centering
\small
\begin{tabular} {  l| l |r| l } 
name & task id & \# instance  & \# feature \\
\hline
car  & {146821} & 1728 & 6 \\
cnae-9 & {9981} & 1080 & 856 \\
connect-4 & {146195} & 67557 & 42 \\
Covertype & {7593} & 581012 & 54 \\
dilbert & {168909} & 10000 & 2000 \\
Dionis & {189355} & 416188 & 60 \\
fabert & {168852} & 8237 & 800 \\
Fashion-MNIST & {146825} & 70000 & 784 \\
Helena & {168329} & 65196 & 27 \\
Jannis & {168330} & 83733 & 54 \\
jungle\_chess\_2pcs... & {167119} & 44819 & 6 \\
mfeat-factors & {12} & 2000 & 216 \\
Robert & {168332} & 10000 & 7200 \\
segment & {146822} & 2310 & 19 \\
shuttle & {146212} & 58000 & 9 \\
vehicle & {53} & 846 & 18 \\
volkert & {168810} & 58310 & 180 \\
 \end{tabular}
 \end{table}

\begin{table}
\caption{Regression Datasets.} \label{tab:regdata}
\centering
\small
\begin{tabular} {  l| l |r| l } 
name & task id & \# instance  & \# feature \\
\hline
2dplanes & {2306} & 40768 & 10 \\
bng\_breastTumor & {7324} & 116640 & 9 \\
bng\_echomonths & {7323} & 17496 & 9 \\
bng\_lowbwt & {7320} & 31104 & 9 \\
bng\_pbc & {7318} & 1000000 & 18 \\
bng\_pharynx & {7322} & 1000000 & 11 \\
bng\_pwLinear & {7325} & 177147 & 10 \\
fried & {4885} & 40768 & 10 \\
house\_16H & {4893} & 22784 & 16 \\
house\_8L & {2309} & 22784 & 8 \\
houses & {5165} & 20640 & 8 \\
mv & {4774} & 40768 & 10 \\
poker & {10102} & 1025010 & 10 \\
pol & {2292} & 15000 & 48 \\
 \end{tabular}
\end{table}

\begin{table}
\caption{\% of tasks where \FLAML\ has better or matching error vs. the baselines when using smaller time budget. }
\centering
\tabcolsep=0.05cm
\begin{tabular} {l c | c| c} 
\FLAML\ vs. Baseline  &  1m vs 10m & 10m vs 1h & 1m vs 1h  \\ 
 \hline
\FLAML\ vs. Auto-sklearn & 65\% & 79\% & 58\%\\
\FLAML\ vs. Cloud-automl &  62\% & 79\% & 48\%\\
\FLAML\ vs. HpBandSter &  63\% & 89\% & 65\%\\
\FLAML\ vs. H2OAutoML & 71\% & 72\% & 50\%\\
\FLAML\ vs. TPOT & 83\% & 85\% & 65\%\\
\end{tabular}
\label{tab:FLAML_res_compare}
\end{table}

\begin{figure*}
    \centering
      \begin{subfigure}{0.33\textwidth}
    \includegraphics[width=\columnwidth]{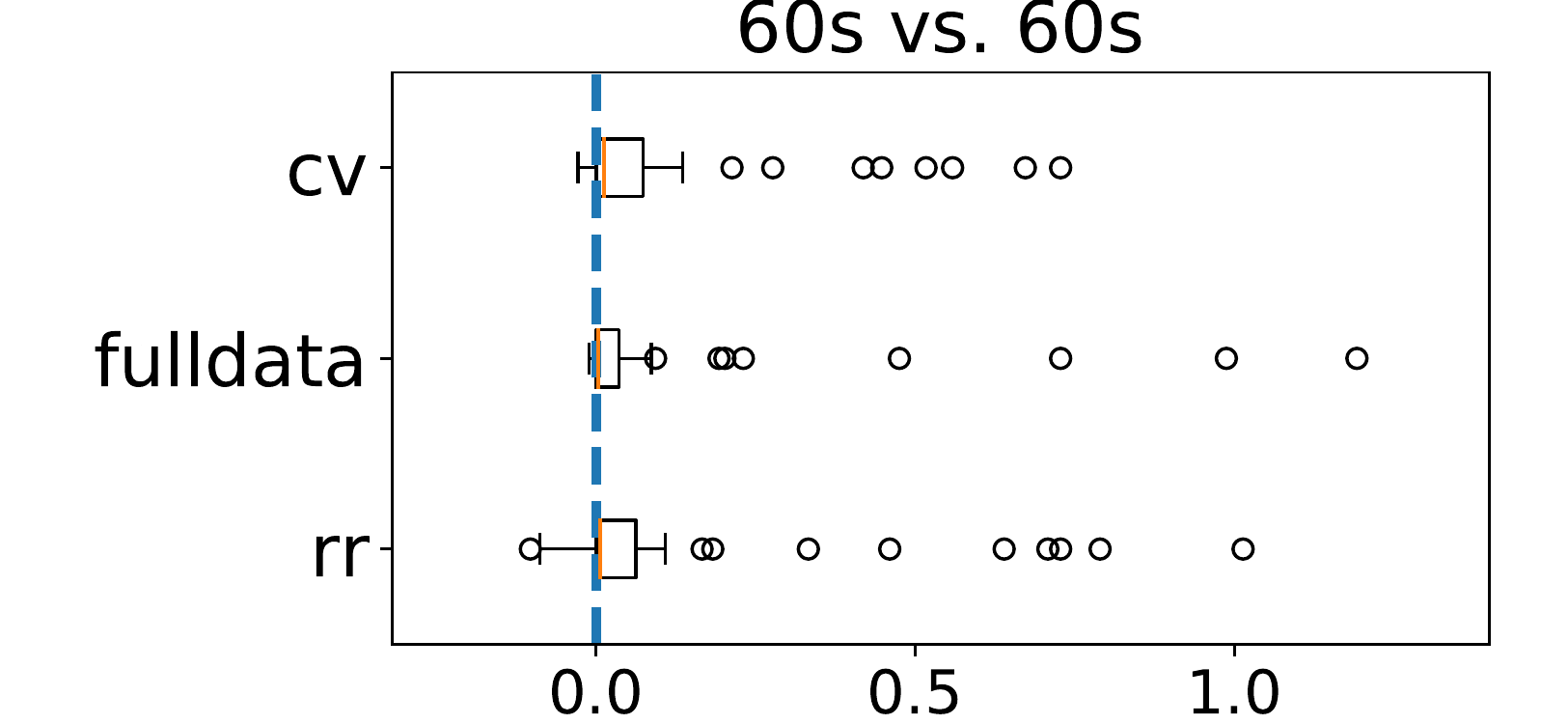}
    \end{subfigure}    
      \begin{subfigure}{0.33\textwidth}
    \includegraphics[width=\columnwidth]{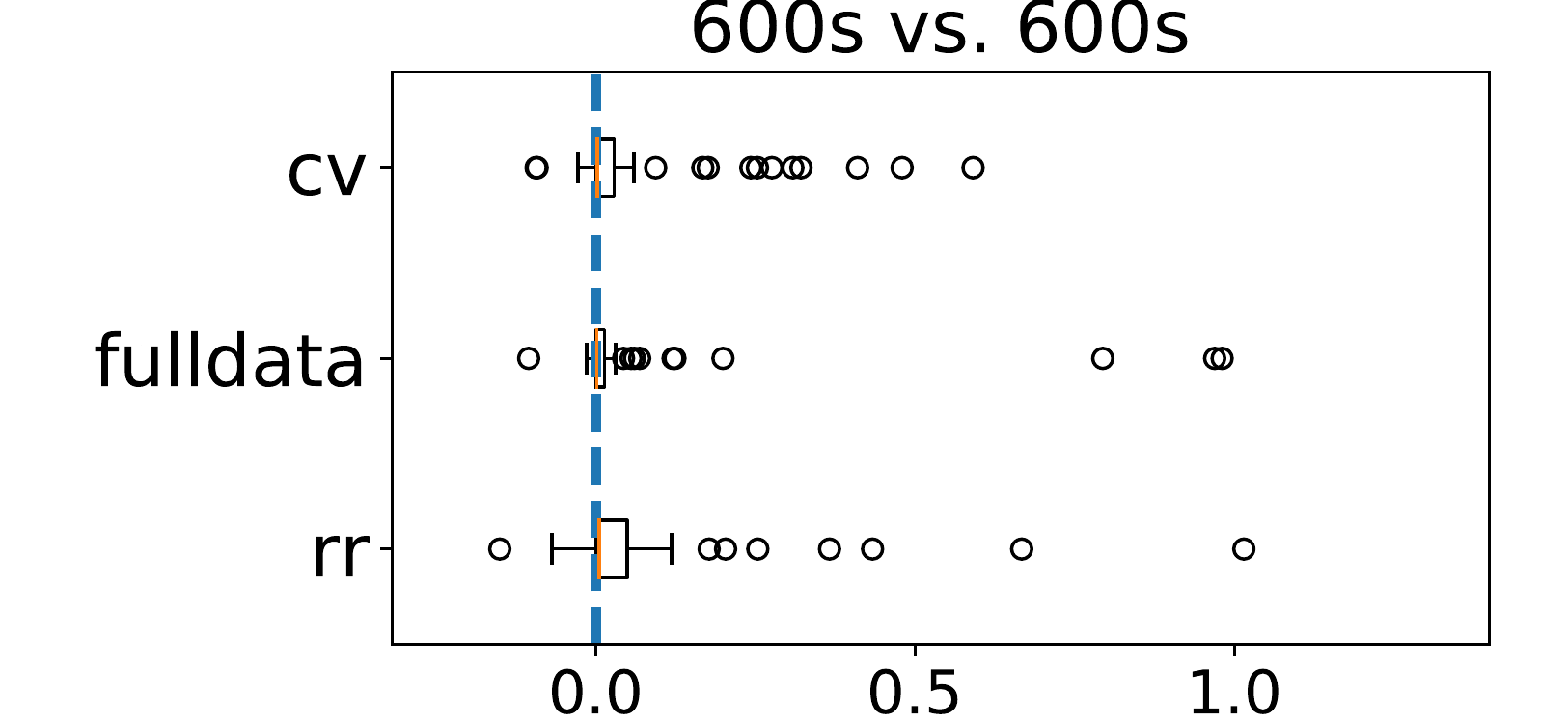}
    \end{subfigure}
      \begin{subfigure}{.33\textwidth}
    \includegraphics[width=\columnwidth]{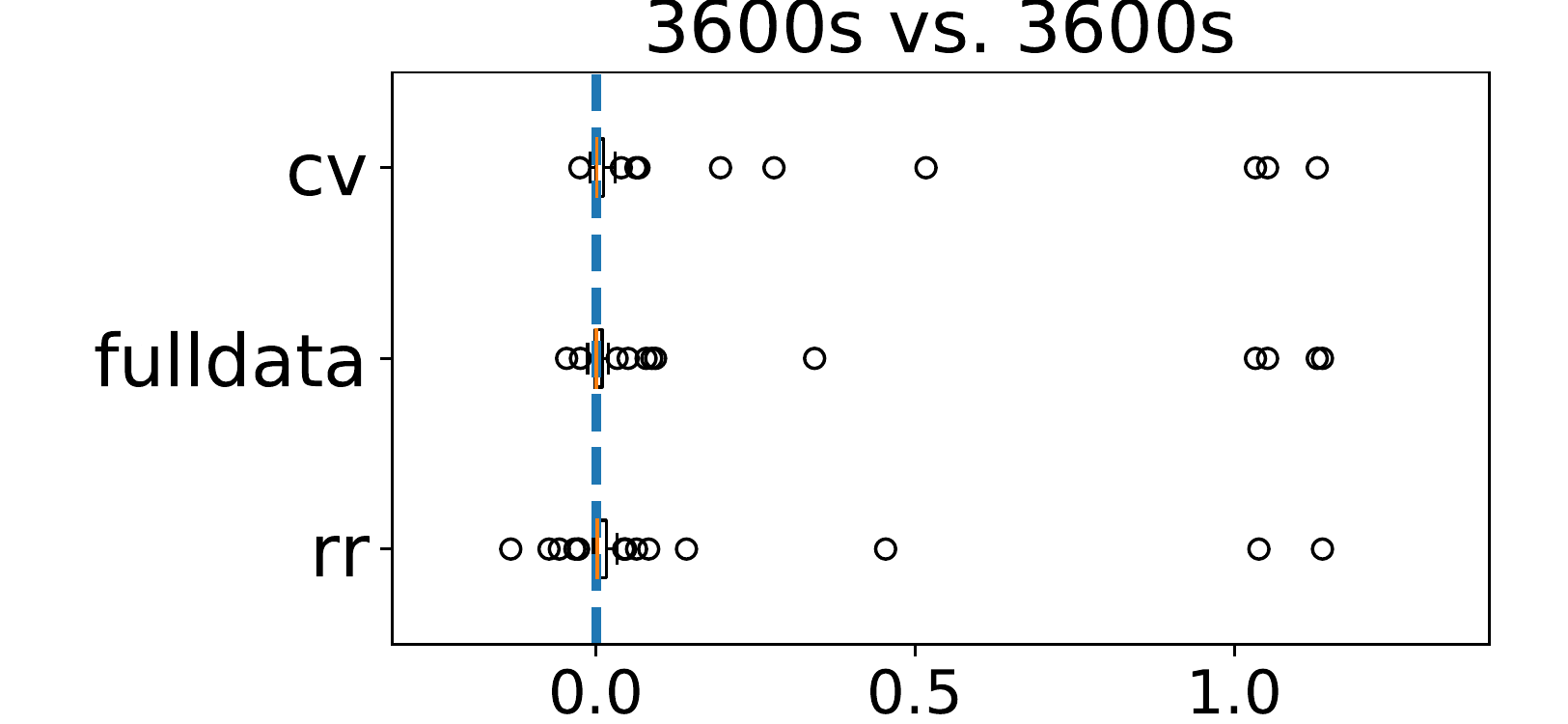}
    \end{subfigure}
    \caption{Score difference for \FLAML\ vs. its own alternatives.}
\label{fig:ablation_box}
\end{figure*}

\section*{Appendix}\label{appendix:system}

The default search space of \FLAML is shown in Table~\ref{tab:searchspace}.

{ For the learners which have not been tried in the search, \(ECI_1(l)\) is set to the smallest trial cost for each learner \(l\). Since the smallest trial cost varies with input data, we first run the fastest learner and get its smallest cost on the input data, and then set the \(ECI_1\) for other learners as multiples of this cost using predefined constants. These constants are easy to set as we only need to calibrate the running time of the fastest configuration of each learner offline.} 
We use the following constants: 
\{`lightgbm':1,
    `xgboost':1.6,
    'extra tree':1.9,
    `rf':2,
    'catboost':15,
    `lr':160
\}. Meta-learning can be potentially applied here to have a more instance-specific prediction of the running time of the initial configuration. We use \(c=2\) as the multiplicative factor of sample size in the experiments.

\FLAML\ is designed to work with low resource consumption, and the extra computation in \FLAML\ beyond trial cost is negligible. 
So it tries one configuration at a time and lets the learner consume all the given resources (cores and RAM). Since we start search from inexpensive models for every learner, this design minimizes the latency between two iterations so that the proposers can get feedback as early as possible. When abundant cores are available and a learner cannot consume all of them, we can extend \FLAML\ to work with multiple search threads in parallel. After choosing one learner based on ECI to perform one search iteration, if there are extra available resources, we can sample another learner by ECI, and so on. When one search iteration for a learner finishes, the resource is released and we can select a learner again using updated ECIs. One learner can also have multiple search threads by using different starting points of the hyperparameters. 
Due to the search space decomposition and the randomized direct hyperparameter search strategy used, the multiple search threads are largely independent and do not interfere with each other.

Stacked ensemble can be added as a post-processing step like  existing libraries~\cite{H2O}. It requires remembering the predictions on cross-validation folds of the models to ensemble. And extra time needs to be spent on building the ensemble and retraining each model. 
\FLAML\ does not do it by default to keep the overhead low, but it offers the option to enable it when storage and extra computation time are not concerns.

Richer types of ML tasks or model assessment criteria can be allowed via customized learners and metric functions. For example, one may search for the cheapest model with error below a threshold using our framework. 

In Table~\ref{tab:FLAML_res_compare}, each row shows the percentage of datasets where \FLAML\ is better than or equal to a particular baseline with smaller time budget. For example, `1m vs 10m' in the header means \FLAML's time budget is one minute and the concerned baseline's time budget is ten minutes. We use a tolerance ratio of 0.1\% to exclude the marginal differences on the scaled scores, i.e., when the difference between two scores is within the tolerance ratio, they are considered as close enough. 
\FLAML's performance in one \textit{minute} is already better than or equal to auto-sklearn, H2O AutoML and TPOT's performance in one \textit{hour} on more than half of the tasks.

%% file: flaml.bbl
\begin{thebibliography}{40}
\providecommand{\natexlab}[1]{#1}
\providecommand{\url}[1]{\texttt{#1}}
\expandafter\ifx\csname urlstyle\endcsname\relax
  \providecommand{\doi}[1]{doi: #1}\else
  \providecommand{\doi}{doi: \begingroup \urlstyle{rm}\Url}\fi

\bibitem[Agrawal et~al.(2020)Agrawal, Chatterjee, Curino, Floratou, Gowdal,
  Interlandi, Jindal, Karanasos, Krishnan, Kroth, et~al.]{agrawal2020cloudy}
Agrawal, A., Chatterjee, R., Curino, C., Floratou, A., Gowdal, N., Interlandi,
  M., Jindal, A., Karanasos, K., Krishnan, S., Kroth, B., et~al.
\newblock Cloudy with high chance of dbms: A 10-year prediction for
  enterprise-grade ml.
\newblock In \emph{CIDR'20}, 2020.

\bibitem[Cormode et~al.(2012)Cormode, Garofalakis, Haas, and
  Jermaine]{Cormode12}
Cormode, G., Garofalakis, M., Haas, P.~J., and Jermaine, C.
\newblock Synopses for massive data: Samples, histograms, wavelets, sketches.
\newblock \emph{Found. Trends Databases}, 4\penalty0 (1–3):\penalty0 1–294,
  January 2012.

\bibitem[Dutt et~al.(2019)Dutt, Wang, Nazi, Kandula, Narasayya, and
  Chaudhuri]{Dutt2019selectivity}
Dutt, A., Wang, C., Nazi, A., Kandula, S., Narasayya, V.~R., and Chaudhuri, S.
\newblock Selectivity estimation for range predicates using lightweight models.
\newblock \emph{{PVLDB}}, 12\penalty0 (9):\penalty0 1044--1057, 2019.

\bibitem[Dutt et~al.(2020)Dutt, Wang, Narasayya, and
  Chaudhuri]{dutt2020efficiently}
Dutt, A., Wang, C., Narasayya, V., and Chaudhuri, S.
\newblock Efficiently approximating selectivity functions using low overhead
  regression models.
\newblock In \emph{46th International Conference on Very Large Data Bases},
  2020.

\bibitem[Elsken et~al.(2019)Elsken, Metzen, and Hutter]{JMLR2019NAS}
Elsken, T., Metzen, J.~H., and Hutter, F.
\newblock Neural architecture search: A survey.
\newblock \emph{Journal of Machine Learning Research}, 20\penalty0
  (55):\penalty0 1--21, 2019.

\bibitem[Erickson et~al.(2020)Erickson, Mueller, Shirkov, Zhang, Larroy, Li,
  and Smola]{agtabular}
Erickson, N., Mueller, J., Shirkov, A., Zhang, H., Larroy, P., Li, M., and
  Smola, A.
\newblock Autogluon-tabular: Robust and accurate automl for structured data.
\newblock \emph{arXiv:2003.06505}, 2020.

\bibitem[Falkner et~al.(2018)Falkner, Klein, and Hutter]{falkner2018}
Falkner, S., Klein, A., and Hutter, F.
\newblock {BOHB}: Robust and efficient hyperparameter optimization at scale.
\newblock In \emph{ICML}, 2018.

\bibitem[Feurer et~al.(2015)Feurer, Klein, Eggensperger, Springenberg, Blum,
  and Hutter]{feurer2015efficient}
Feurer, M., Klein, A., Eggensperger, K., Springenberg, J., Blum, M., and
  Hutter, F.
\newblock Efficient and robust automated machine learning.
\newblock In \emph{NIPS}, 2015.

\bibitem[Feurer et~al.(2020)Feurer, Eggensperger, Falkner, Lindauer, and
  Hutter]{ASKL2}
Feurer, M., Eggensperger, K., Falkner, S., Lindauer, M., and Hutter, F.
\newblock Auto-sklearn 2.0.
\newblock \emph{arXiv:2007.04074}, 2020.

\bibitem[Fusi et~al.(2018)Fusi, Sheth, and Elibol]{fusi2018probabilistic}
Fusi, N., Sheth, R., and Elibol, M.
\newblock Probabilistic matrix factorization for automated machine learning.
\newblock In \emph{Advances in Neural Information Processing Systems}, 2018.

\bibitem[Galakatos et~al.(2019)Galakatos, Markovitch, Binnig, Fonseca, and
  Kraska]{galakatos2019tree}
Galakatos, A., Markovitch, M., Binnig, C., Fonseca, R., and Kraska, T.
\newblock Fiting-tree: A data-aware index structure.
\newblock In \emph{SIGMOD}, 2019.

\bibitem[Gijsbers et~al.(2019)Gijsbers, LeDell, Thomas, Poirier, Bischl, and
  Vanschoren]{Gijsbers2019benchmark}
Gijsbers, P., LeDell, E., Thomas, J., Poirier, S., Bischl, B., and Vanschoren,
  J.
\newblock An open source automl benchmark.
\newblock In \emph{AutoML Workshop at ICML 2019}, 2019.
\newblock URL \url{http://arxiv.org/abs/1907.00909}.

\bibitem[H2O.ai()]{H2O}
H2O.ai.
\newblock H2o automl.
\newblock \url{http://docs.h2o.ai/h2o/latest-stable/h2o-docs/automl.html}.
\newblock 2019-10-29.

\bibitem[Hastie et~al.(2001)Hastie, Tibshirani, and
  Friedman]{hastie01statisticallearning}
Hastie, T., Tibshirani, R., and Friedman, J.
\newblock \emph{The Elements of Statistical Learning}.
\newblock Springer New York Inc., New York, NY, USA, 2001.

\bibitem[Horn et~al.(2019)Horn, Pack, and Rieger]{horn2019autofeat}
Horn, F., Pack, R., and Rieger, M.
\newblock The autofeat python library for automatic feature engineering and
  selection.
\newblock In \emph{Machine Learning and Knowledge Discovery in Databases. ECML
  PKDD 2019}, 2019.

\bibitem[Huang et~al.(2019)Huang, Wang, Ding, and
  Chaudhuri]{huang2019efficient}
Huang, S., Wang, C., Ding, B., and Chaudhuri, S.
\newblock Efficient identification of approximate best configuration of
  training in large datasets.
\newblock In \emph{Proceedings of the 33rd AAAI Conference on Artificial
  Intelligence (AAAI)}, 2019.

\bibitem[Hutter et~al.(2011)Hutter, Hoos, and Leyton-Brown]{hutter2011}
Hutter, F., Hoos, H.~H., and Leyton-Brown, K.
\newblock Sequential model-based optimization for general algorithm
  configuration.
\newblock In Coello, C. A.~C. (ed.), \emph{Learning and Intelligent
  Optimization}, 2011.

\bibitem[Kipf et~al.(2019)Kipf, Kipf, Radke, Leis, Boncz, and
  Kemper]{kipf2019learned}
Kipf, A., Kipf, T., Radke, B., Leis, V., Boncz, P., and Kemper, A.
\newblock Learned cardinalities: Estimating correlated joins with deep
  learning.
\newblock In \emph{CIDR}, 2019.

\bibitem[Kohavi(1995)]{Kohavi1995}
Kohavi, R.
\newblock A study of cross-validation and bootstrap for accuracy estimation and
  model selection.
\newblock In \emph{IJCAI’95}, 1995.

\bibitem[Kraska et~al.(2018)Kraska, Beutel, Chi, Dean, and
  Polyzotis]{kraska2018case}
Kraska, T., Beutel, A., Chi, E.~H., Dean, J., and Polyzotis, N.
\newblock The case for learned index structures.
\newblock In \emph{SIGMOD}, 2018.

\bibitem[Li et~al.(2017)Li, Jamieson, DeSalvo, Rostamizadeh, and
  Talwalkar]{ICLR:li2017hyperband}
Li, L., Jamieson, K., DeSalvo, G., Rostamizadeh, A., and Talwalkar, A.
\newblock Hyperband: A novel bandit-based approach to hyperparameter
  optimization.
\newblock In \emph{ICLR'17}, 2017.

\bibitem[Li et~al.(2020)Li, Jamieson, Rostamizadeh, Gonina, Ben-tzur, Hardt,
  Recht, and Talwalkar]{Li2020}
Li, L., Jamieson, K., Rostamizadeh, A., Gonina, E., Ben-tzur, J., Hardt, M.,
  Recht, B., and Talwalkar, A.
\newblock A system for massively parallel hyperparameter tuning.
\newblock In \emph{Proceedings of Machine Learning and Systems}, 2020.

\bibitem[Liaw et~al.(2019)Liaw, Bhardwaj, Dunlap, Zou, Gonzalez, Stoica, and
  Tumanov]{Liaw19}
Liaw, R., Bhardwaj, R., Dunlap, L., Zou, Y., Gonzalez, J.~E., Stoica, I., and
  Tumanov, A.
\newblock Hypersched: Dynamic resource reallocation for model development on a
  deadline.
\newblock SoCC'19, 2019.

\bibitem[Liberty et~al.(2020)Liberty, Karnin, Xiang, Rouesnel, Coskun,
  Nallapati, Delgado, Sadoughi, Astashonok, Das, et~al.]{liberty2020elastic}
Liberty, E., Karnin, Z., Xiang, B., Rouesnel, L., Coskun, B., Nallapati, R.,
  Delgado, J., Sadoughi, A., Astashonok, Y., Das, P., et~al.
\newblock Elastic machine learning algorithms in amazon sagemaker.
\newblock In \emph{SIGMOD’20}, 2020.

\bibitem[Ma et~al.(2018)Ma, Aken, Hefny, Mezerhane, Pavlo, and
  Gordon]{ma2018query}
Ma, L., Aken, D.~V., Hefny, A., Mezerhane, G., Pavlo, A., and Gordon, G.~J.
\newblock Query-based workload forecasting for self-driving database management
  systems.
\newblock In \emph{SIGMOD}, 2018.

\bibitem[Marcus \& Papaemmanouil(2019)Marcus and
  Papaemmanouil]{pvldb:MarcusP19}
Marcus, R.~C. and Papaemmanouil, O.
\newblock Plan-structured deep neural network models for query performance
  prediction.
\newblock \emph{{PVLDB}}, 12\penalty0 (11):\penalty0 1733--1746, 2019.

\bibitem[Mukunthu et~al.(2019)Mukunthu, Shah, and Tok]{mukunthu2019practical}
Mukunthu, D., Shah, P., and Tok, W.
\newblock \emph{Practical Automated Machine Learning on Azure: Using Azure
  Machine Learning to Quickly Build AI Solutions}.
\newblock O'Reilly Media, Incorporated, 2019.

\bibitem[Nakkiran et~al.(2020)Nakkiran, Kaplun, Bansal, Yang, Barak, and
  Sutskever]{nakkiran2019deep}
Nakkiran, P., Kaplun, G., Bansal, Y., Yang, T., Barak, B., and Sutskever, I.
\newblock Deep double descent: Where bigger models and more data hurt.
\newblock In \emph{ICLR}, 2020.

\bibitem[Neumann \& Freitag(2020)Neumann and Freitag]{Neumann20umbra}
Neumann, T. and Freitag, M.~J.
\newblock Umbra: {A} disk-based system with in-memory performance.
\newblock In \emph{CIDR}, 2020.

\bibitem[Olson et~al.(2016)Olson, Urbanowicz, Andrews, Lavender, Kidd, and
  Moore]{Olson2016TPOT}
Olson, R.~S., Urbanowicz, R.~J., Andrews, P.~C., Lavender, N.~A., Kidd, L.~C.,
  and Moore, J.~H.
\newblock Automating biomedical data science through tree-based pipeline
  optimization.
\newblock In Squillero, G. and Burelli, P. (eds.), \emph{Applications of
  Evolutionary Computation}, pp.\  123--137. Springer International Publishing,
  2016.

\bibitem[Olson et~al.(2017)Olson, La~Cava, Orzechowski, Urbanowicz, and
  Moore]{Olson2017PMLB}
Olson, R.~S., La~Cava, W., Orzechowski, P., Urbanowicz, R.~J., and Moore, J.~H.
\newblock Pmlb: a large benchmark suite for machine learning evaluation and
  comparison.
\newblock \emph{BioData Mining}, 10\penalty0 (1):\penalty0 36, Dec 2017.
\newblock ISSN 1756-0381.
\newblock \doi{10.1186/s13040-017-0154-4}.
\newblock URL \url{https://doi.org/10.1186/s13040-017-0154-4}.

\bibitem[Oracle docs()]{oracle_stats}
Oracle docs.
\newblock Optimizer statistics (release 18).
\newblock
  \url{https://docs.oracle.com/en/database/oracle/oracle-database/18/tgsql/optimizer-statistics.html\#GUID-0A2F3D52-A135-43E1-9CAB-55BFE068A297}.

\bibitem[Pedregosa et~al.(2011)Pedregosa, Varoquaux, Gramfort, Michel, Thirion,
  Grisel, Blondel, Prettenhofer, Weiss, Dubourg, Vanderplas, Passos,
  Cournapeau, Brucher, Perrot, and Duchesnay]{JMLR:Pedregosa2011}
Pedregosa, F., Varoquaux, G., Gramfort, A., Michel, V., Thirion, B., Grisel,
  O., Blondel, M., Prettenhofer, P., Weiss, R., Dubourg, V., Vanderplas, J.,
  Passos, A., Cournapeau, D., Brucher, M., Perrot, M., and Duchesnay, E.
\newblock Scikit-learn: Machine learning in python.
\newblock \emph{JMLR}, 12:\penalty0 2825--2830, November 2011.

\bibitem[Renggli et~al.(2019)Renggli, Karlaš, Ding, Liu, Wu, and
  Zhang]{renggli2019continuous}
Renggli, C., Karlaš, B., Ding, B., Liu, F., Wu, W., and Zhang, C.
\newblock Continuous integration of machine learning models with ease.ml/ci:
  Towards a rigorous yet practical treatment.
\newblock In \emph{SysML Conference (SysML 2019)}, 2019.

\bibitem[Selinger et~al.(1979)Selinger, Astrahan, Chamberlin, Lorie, and
  Price]{Selinger79}
Selinger, P.~G., Astrahan, M.~M., Chamberlin, D.~D., Lorie, R.~A., and Price,
  T.~G.
\newblock Access path selection in a relational database management system.
\newblock SIGMOD'79, 1979.

\bibitem[Shang et~al.(2019)Shang, Zgraggen, Buratti, Kossmann, Eichmann, Chung,
  Binnig, Upfal, and Kraska]{Shang2019DDS}
Shang, Z., Zgraggen, E., Buratti, B., Kossmann, F., Eichmann, P., Chung, Y.,
  Binnig, C., Upfal, E., and Kraska, T.
\newblock Democratizing data science through interactive curation of ml
  pipelines.
\newblock In \emph{SIGMOD}, 2019.

\bibitem[Snoek et~al.(2012)Snoek, Larochelle, and Adams]{snoek2012practical}
Snoek, J., Larochelle, H., and Adams, R.~P.
\newblock Practical bayesian optimization of machine learning algorithms.
\newblock In \emph{NIPS}, 2012.

\bibitem[SQL Server docs()]{sqlserver_stats}
SQL Server docs.
\newblock Statistics in \uppercase{M}icrosoft \uppercase{SQL S}erver 2017.
\newblock
  \url{https://docs.microsoft.com/en-us/sql/relational-databases/statistics/statistics?view=sql-server-2017}.

\bibitem[Vanschoren et~al.(2014)Vanschoren, van Rijn, Bischl, and
  Torgo]{Vanschoren2014}
Vanschoren, J., van Rijn, J.~N., Bischl, B., and Torgo, L.
\newblock Openml: Networked science in machine learning.
\newblock \emph{SIGKDD Explor. Newsl.}, 15\penalty0 (2):\penalty0 49–60, June
  2014.

\bibitem[Wu et~al.(2021)Wu, Wang, and Huang]{wu2021cost}
Wu, Q., Wang, C., and Huang, S.
\newblock Frugal optimization for cost-related hyperparameters.
\newblock In \emph{AAAI'21}, 2021.

\end{thebibliography}
